\documentclass[11pt]{article}

\usepackage[final]{emnlp2023}

\usepackage{times}
\usepackage{latexsym}

\usepackage{amssymb}

\usepackage[T1]{fontenc}

\usepackage[utf8]{inputenc}

\usepackage[most]{tcolorbox}
\usepackage{enumitem}

\definecolor{warmpeach}{RGB}{255, 228, 196}
\definecolor{warmborder}{RGB}{210, 105, 30}

\usepackage{microtype}

\usepackage{inconsolata}
\usepackage{booktabs}

\usepackage{graphicx}

\usepackage{listings}
\usepackage[most]{tcolorbox}
\usepackage{multirow}
\usepackage{subcaption}
\usepackage{float}

\usepackage{eso-pic}
\usepackage{pifont}
\usepackage{array}
\newcolumntype{H}{>{\setbox0=\hbox\bgroup}c<{\egroup}@{}}

\definecolor{promptbg}{RGB}{245,245,250}
\definecolor{promptframe}{RGB}{160,160,180}

\newtcolorbox{promptbox}[1][]{%
  enhanced,
  breakable,
  colback=promptbg,
  colframe=promptframe,
  boxrule=0.5pt,
  arc=2pt,
  left=6pt, right=6pt, top=4pt, bottom=4pt,
  fontupper=\small\ttfamily,
  title={#1},
  coltitle=black,
  fonttitle=\small\bfseries\sffamily,
  attach boxed title to top left={yshift=-2mm, xshift=4mm},
  boxed title style={
    colback=promptbg,
    colframe=promptframe,
    boxrule=0.5pt,
    arc=1pt,
    left=3pt, right=3pt, top=1pt, bottom=1pt,
  },
}

\newcommand{\vars}[1]{\par\vspace{4pt}\noindent{\small\sffamily\textbf{Variables:} #1}}
\newcommand\cut[1]{}
\title{Rethinking Molecular Text Representations for LLMs:\\ An Empirical Study}

\author{Arun Raja \thanks{~~\texttt{arun.raja@dtc.ox.ac.uk}} \\
  University of Oxford\\
   \\\And
  Garrett M. Morris \\
  University of Oxford \\
\\\And
  Kian Ming A. Chai \\
  DSO National Laboratories \\
}

\begin{document}

\maketitle
\begin{abstract}

Large language models (LLMs) are increasingly used for molecular tasks, but it remains unclear \emph{which molecular representation to use}. We present a systematic benchmark, MolRepBench, to evaluate LLM molecular competence across nine representations and eight chemical tasks. We benchmark 16 LLMs across five model families, including reasoning and non-reasoning variants, chemistry-specialized LLMs, and closed frontier models. Performance is strongly representation-dependent and no single representation wins across tasks, though CML is the best, followed by MolJSON, InChI and canonical SMILES. Explicit structured text representations (CML and MolJSON) dominate structural tasks; IUPAC dominates semantic tasks, winning molecule retrieval for all 16 LLMs; and SMILES variants are rarely optimal despite their prevalence in pretraining. Chemistry-specialized models perform well with SMILES at the cost of large degradations with structured text representations, suggesting SMILES-only evaluation rewards specialization that does not generalize. Using LLM-as-a-judge, we find that IUPAC produces the highest fraction of correct molecule generations. A mechanistic study via tokenization audits, linear probes, and attention shows that representations are encoded differently inside the model; for example, structured representations require more attention across the molecular span. Our results argue against representation-invariant evaluation and motivate task-aware representation routing for LLM-based chemistry.

\end{abstract}

\section{Introduction}

\label{sec:intro}

When using large language models (LLMs) for chemistry, molecules can be expressed in various text-based formats. However, the representation used for a task is mostly chosen by convention rather than empirical evidence.
A medicinal chemist querying an LLM might give a SMILES string, an IUPAC name, or a SELFIES string---yet we have no systematic understanding of which representation best supports different tasks.

Prior work on molecular LLMs focuses either on training models from scratch on molecular strings \citep{edwards-etal-2022-translation,liu2023moleculestm} or on finetuning foundation models for specific tasks \citep{zeng2022deep,zhao2023gimlet}. Most such works adopt SMILES \citep{weininger1988smiles} as the de facto representation, given its prevalence in chemistry literature and applications. However, there are molecular text representations for different purposes.

Chemistry has diverse machine-readable representations, each different in compactness, invertibility, and application \citep{ az_rev,wiley_rev, wiley_rev2}. An early textual representation was the Wiswesser Line Notation (WLN, \citealt{wln}). It was designed to be punched onto cards and processed by tabulating machines, and it was the dominant machine-readable representation in the industry.

\citet{iupac1979} introduced a nomenclature for the International Union of Pure and Applied Chemistry (IUPAC). IUPAC maintains rules to name any molecule in natural language. It is more human-readable than other text notations. However, the rules are elaborate with edge cases, limiting scalability. In contrast to WLN and IUPAC, Simplified Molecular Input Line Entry System
 (SMILES, \citealt{weininger1988smiles}) is a context-free grammar and can be algorithmically derived. It was created for modern computer software to process and store chemical structures efficiently. Its canonical variant, canonical SMILES, ensured uniqueness for each molecule, so it could be used as a database key. The advent of generative AI in chemistry introduced DeepSMILES \citep{o2018deepsmiles} and SELFIES \citep{krenn2020selfies}, designed to reduce invalid generations from probabilistic models. 
 
To store, access, and exchange descriptions of molecules on the internet, Extensible Markup Language and JavaScript Object Notation 
have become popular. 
Chemical Markup Language (CML, \citealt{cml2000, cml2003}) is the first such; 
MolJSON \citep{moljson} is the latest (see Section~\ref{appendix_structured}).
They are called \textbf{structured text representations} because they explicitly store graph information such as atoms and bond connectivity.

Therefore, there are dominant molecular representations for different purposes and technology.
Presently, LLM is a prevalent technology. It has its own affordances (tokenization, in-context learning, alignment with natural language) and failure modes (for example, hallucination).
\emph{Which representation should chemistry use for an LLM?}

We address the question via a thorough, systematic study of various molecular text representations for use with LLMs.
Our contributions are:

\begin{enumerate}
    \item \textbf{MolRepBench}, a benchmark of eight tasks covering comprehension and generation, each prompted with nine representation formats.
    \item \textbf{Evaluation across five model families}, spanning non-reasoning and reasoning-capable models, domain-specialized models and closed frontier models. We use \textbf{LLM-as-a-judge} to qualitatively evaluate the molecules generated with different representations.
    \item \textbf{A mechanistic interpretability study} in which we investigate how an LLM `views' the various molecular representations.
\end{enumerate}

Our findings conclude that current LLMs do not yet possess robust, representation-invariant molecular capabilities. Thus, molecular representation is a key factor when using LLMs for chemistry.

\section{Related Work}
\label{sec:related}

\textbf{The effect of representations on LLM performance} has been explored in general. For instance, \citet{yauney-mimno-2021-comparing} show that the difficulty of a dataset is related to the alignment between the input
representation and the labeling. \citet{al-shaibani-ahmad-2023-consonant} suggest a consonant-based compact representation for English using less computational resources yet achieving performance comparable to the standard text representation. Our work uses both compact and verbose representations to investigate how verbosity impacts an LLM's performance across different chemistry tasks.

A growing body of work applies \textbf{large language models (LLMs) to chemistry}. Domain-specific models such as MolT5 \citep{edwards-etal-2022-translation}, Text+Chem T5 \citep{christofidellis2023unifying}, Galactica \citep{taylor2022galacticalargelanguagemodel} and nach0 \citep{nach0} jointly model SMILES and natural language for molecule captioning, caption-to-molecule generation, and property prediction. General-purpose LLMs have also been evaluated on chemistry tasks through prompting and in-context learning \citep{Jablonka2024LeveragingLL, Li_2024}, and recent benchmarks such as ChemLLMBench  \citep{guo_bench} and ChemBench \citep{mirza2024large} measure performance across property prediction, name conversion, and structural reasoning. However, the dominant representation across this literature is SMILES; there are only occasional comparisons to SELFIES or to text-encoded molecular graphs.

 
 A \textbf{prior work comparing representations for molecules} by \citet{Baker2025MolecularSR} studies canonical SMILES, DeepSMILES, SELFIES, InChI and IUPAC names for molecular property prediction.
Our work builds upon this in \emph{three} aspects (see Table~\ref{tab:prior-work}).
First, the prior work has not explored 
structured text representations, which may be important given that graph neural networks (GNNs) outperform LLMs in molecular property prediction \citep{Zhong2024BenchmarkingLL, Gupta_2025, jacobs2026regressionlargelanguagemodels}.
Our work evaluates CML and MolJSON, which textually encode the molecular graph.

Given that LLMs are known to be less competent than GNNs for property \emph{prediction}, comparing text representations on predicting properties, such as solubility, toxicity, and lipophilicity, may offer limited insight. These biochemical properties are measured using wet-lab assays, and their reliable predictions require chemical or physical inductive biases. Therefore, we instead frame such tasks as molecular property \emph{estimation}, where the LLMs are to predict the number of hydrogen bond donors and acceptors,  Wildman-Crippen LogP value and topological polar surface area or TPSA for molecules. 
These properties are computable from the molecular graph (using RDKit, \citealt{rdkit}) without wet-lab experiments.
This is the second aspect.

The third aspect is data contamination. \citet{Baker2025MolecularSR} uses datasets from the widely used MoleculeNet \citep{moleculenet}. These datasets are likely in the pretraining corpus \citep{sainz-etal-2023-nlp, cheng2025surveydatacontaminationlarge,xu2025reimaginesymbolicbenchmarksynthesis, blinding}. 
By contrast, our property labels are generated using RDKit, so are unavailable elsewhere.

\begin{table*}[t]
\centering
\small
\renewcommand{\arraystretch}{1.3}
\begin{tabular}{@{}p{0.12\linewidth}p{0.40\linewidth}p{0.42\linewidth}@{}}
\toprule
\textbf{Aspect} & \textbf{Prior work }\citep{Baker2025MolecularSR} & \textbf{Ours} \\
\midrule
Representations
& Canonical SMILES, DeepSMILES, SELFIES, InChI, IUPAC. \emph{No} structured text respresentations.
& In addition to Baker et al.'s, isomeric SMILES, randomized SMILES, \textbf{CML} and \textbf{MolJSON}. The last two are structured text representations. \\
\addlinespace
Task framing
& Molecular \emph{property prediction} (e.g., solubility, toxicity, lipophilicity): wet-lab properties requiring chemical/physical inductive biases that LLMs lack
& Molecular \emph{property estimation} (e.g., HBD, HBA, Wildman--Crippen LogP, TPSA): properties computable from molecular graphs without wet-lab assays \\
\addlinespace
Label in Data
& MoleculeNet datasets, likely present in LLM pretraining corpora
& Labels generated with RDKit, unavailable elsewhere, mitigating contamination \\
\bottomrule
\end{tabular}
\caption{Comparison of MolRepBench with the prior representation-comparison study by \citet{Baker2025MolecularSR}.}
\label{tab:prior-work}
\end{table*}

\section{Benchmark Design}
\label{sec:benchmark}

We evaluate \textbf{ nine molecular representations} (Table~\ref{tab:repr_examples_lala} and Appendix~\ref{fig:appendix_repr_examples}).
Each representation encodes the same molecule but differs in syntactic structure and implicit assumptions. We decided on these nine based on their prevalent use in modern cheminformatics tools and deep learning methods. SMILES is the de facto standard and is widely represented in LLM pretraining corpora. We include three variants: canonical SMILES determines the atom ordering algorithmically; isomeric SMILES adds stereochemical annotations; and randomized SMILES removes canonical ordering by generating a new traversal each time. 

Next, we consider two representations developed for generative models. First, DeepSMILES is a derivative of SMILES that eliminates the two grammatical features most prone to autoregressive failure:  matched parentheses and paired ring-closure digits. Second, SELFIES guarantees that every string decodes into a valid molecule. 
We evaluate whether such syntactic fixes help LLMs. 

We also study two notations from IUPAC, namely IUPAC name and International Chemical Identifier (InChI, \citealt{inchi_mcnaught2006iupac}). IUPAC names are linguistically compositional and human-readable, so the LLM can parse it as natural language rather than as symbolic code.
They are also ubiquitous in chemistry texts and chemistry pretraining corpora \citep{Baker2025MolecularSR}.
InChI is a compact string with detailed information, such as the chemical formula, atom connections, charge and stereochemistry, organized in layers and sub-layers. 
InChI strings are also canonical. 

Lastly, we have CML and MolJSON as structured representations with adjacency information.

\begin{table*}[t]
\centering
\footnotesize
\begin{tabular}{@{}l@{ }lp{6.30cm}p{5.12cm}@{}}
\toprule
\textbf{Representation}& \textbf{Year} & \textbf{Description} & \textbf{Example for L-Alanine (C$_3$H$_7$NO$_2$)} \\
\midrule
Canonical SMILES  & 1988 & Short ASCII strings with canonical ordering & \texttt{CC(N)C(=O)O} \\
Isomeric SMILES   & 1988 & SMILES with isotopic and chiral specifications & \texttt{C[C@@H](N)C(=O)O} \\
Randomized SMILES & 2017 & Non-canonical atom ordering, generated fresh per call using RDKit & \texttt{C(C(=O)O)(N)C} \\
DeepSMILES        & 2018 & Removes matching parentheses and ring-closure digits & \texttt{CCN)C=O)O} \\
IUPAC name        & 1979 & Natural language description that follows IUPAC nomenclature & \texttt{(2S)-2-aminopropanoic acid} \\
SELFIES           & 2020 & Self-referencing embedded strings & \texttt{[C][C@@H1][Branch1]...[=O][O][N]} \\
CML               & 2000 & XML-based explicit graph & \texttt{<molecule><atomArray>... </atomArray><bondArray>... </bondArray></molecule>}  \raggedright\tabularnewline
InChI             & 2005 & Layered canonical identifier by IUPAC & \texttt{InChI=1S/C3H7NO2/c1-2(4)...} \raggedright\tabularnewline
MolJSON           & 2026 & Explicit graph representation in JSON & \texttt{\{"atoms": [...],"bonds": [...]\}} \raggedright\tabularnewline
\bottomrule
\end{tabular}

\caption{The nine molecular representations used in this work, with the year of introduction. For the L-Alanine example, SELFIES, CML, InChI, and MolJSON are abbreviated. Complete forms are provided in Section \ref{fig:appendix_repr_examples}.
}

\label{tab:repr_examples_lala}
\end{table*}

Our benchmark, \textbf{MolRepBench}, has 8 molecular reasoning and generation tasks, spanning low-level structural parsing, chemically grounded semantic recognition, retrieval, and caption-to-molecule generation.
\footnote{The code for MolRepBench is available at \url{https://github.com/arunraja-hub/molrepbench}.}
It draws molecular data from ChEBI-20 \citep{edwards-etal-2022-translation} and ZINC250K \citep{zinc}. 
The tasks are summarised below (and in Table~\ref{tab:tasks}; details in Sections~\ref{source_datasets} and~\ref{appendix_data_tasks}).

\begin{description}[style=unboxed,leftmargin=0.5cm,noitemsep]
    \item[Atom Counting] tests whether the LLM can accurately parse molecular strings and count atoms of specified elements, probing low-level structural comprehension.
    
    \item[Functional Group Identification] evaluates the LLM's ability to recognize chemically meaningful substructures (aldehyde, ester, halide, primary amine, sulfonamide).
    
    \item[Molecular Property Estimation] assesses the prediction of physicochemical properties---LogP, TPSA, hydrogen-bond donors (HBD) and hydrogen-bond acceptors (HBA)---directly from the representations, testing quantitative chemical reasoning.
    
    \item[Molecule Retrieval] measures alignment between natural language descriptions and molecular structures using constructed distractors (similar scaffold, similar weight, and random ones).
    
    \item[Isomer Discrimination] probes whether the LLM can distinguish molecules that differ subtly in atom ordering or stereochemistry. This tests fine-grained structural sensitivity.
    
    \item[Tautomer Recognition] evaluates the LLM's understanding of dynamic structural equivalence, where two molecules are the same under relocation of a hydrogen atom or bond rearrangement.
    
    \item[Protonation State Recognition] tests whether the LLM can identify molecules that are chemically identical but differ in protonation, assessing awareness of charge-state equivalence.
    
    \item[Caption-to-Molecule Generation] examines the LLM's generative capability to translate natural language descriptions into valid molecular strings, evaluating both syntactic validity and semantic fidelity.
\end{description}


We employ \textbf{various metrics} (Table \ref{tab:tasks}) to evaluate across our diverse tasks. Atom counting uses exact-match accuracy. Functional group identification uses macro-F1 for the 5 binary labels, one for each functional group. In property estimation, logP and TPSA are regression tasks using Spearman correlation $\rho$, while hydrogen-bond donor/acceptor counts use exact-match accuracy. Molecule retrieval is a multiple-choice task scored by top-1 accuracy. Isomer discrimination, tautomer recognition, and protonation-state recognition are binary classification tasks using accuracy. For caption-to-molecule generation, we report validity rate, exact-match accuracy, Tanimoto similarities \citep{tanimoto1957ibm} on Morgan \citep{morgan}, MACCS \citep{maccs} and RDKit fingerprints \citep{rdkit}, and Fréchet ChemNet Distance \citep{fcd}.

To further evaluate the caption-to-molecule generation task, we also use a closed frontier LLM, Gemini 3 Flash \citep{google2025gemini3flash}, as an \textbf{LLM-as-a-judge} to evaluate the generations. This provides qualitative insights into how each model and representation leads to various failure modes, such as hallucinations.

\begin{table}[t]
\centering
\small
\begin{tabular}{@{}l@{}llp{2.3cm}@{}}
\toprule
& \textbf{Tasks} & \textbf{Sources} & \textbf{Metrics} \\
\midrule
\makebox[1.5ex][l]{Comprehension}\\
&Atom Counting         & ChEBI-20   & Exact match \\
&Group identification       & ChEBI-20   & Macro-F1 \\ 
&Property estimation  & ChEBI-20  & $\rho$, Exact match \\
&Molecule Retrieval    & ChEBI-20   & Top-1 \\
[1.5ex]
\makebox[1.5ex][l]{Discrimination}\\
&Isomer Discrimination       & ChEBI-20   & Accuracy \\
&Tautomer Recognition       & ZINC250K   & Accuracy \\
&Protonation State     & ZINC250K   & Accuracy \\
[1.5ex]
\makebox[1.5ex][l]{Generation}\\
&Caption-to-Molecule  & ChEBI-20   & Validity rate, \newline Tanimoto sim., \newline Exact match, FCD \raggedright \tabularnewline 
\bottomrule
\end{tabular}
\caption{Overview of eight benchmark tasks developed to study how different text representations affect the performance of LLMs in chemistry.}
\label{tab:tasks}
\end{table}


\section{Experiments}
\label{sec:models}

We use 16 large language models (LLMs) spanning five families: three open-weight model families covering a range of sizes and reasoning capabilities, three LLMs specialized for chemistry (and their corresponding base models), and two closed frontier models (Table~\ref{tab:models}; details in Section \ref{modelfamilies}).


\begin{table}[t]
\centering
\small
\begin{tabular}{@{}lllHHc@{}}
\toprule
\textbf{Family} & \textbf{\#Param} & \textbf{Model} & \textbf{Family} & \textbf{Parameters} & \textbf{R} \\
\midrule
Qwen3
& 4B 
 & Qwen3-4B              & \multirow{4}{*}{Qwen3}   & \multirow{2}{*}{4B}                & \ding{51} \\
&& Qwen3-4B              &                          &                                     & \ding{55} \\
&30B 
 & Qwen3-30B-A3B         &                          & \multirow{2}{*}{30B (3B active)}   & \ding{51} \\
&& Qwen3-30B-A3B         &                          &                                     & \ding{55} \\
[1.5ex]
Phi-4
& 14B
 & Phi-4                 & \multirow{3}{*}{Phi-4}   & \multirow{3}{*}{14B}               & \ding{51} \\
&& Phi-4-Reasoning       &                          &                                     & \ding{51} \\
&& Phi-4-Reasoning-Plus  &                          &                                     & \ding{51} \\
[1.5ex]
OLMo
& 32B
 & OLMo-3.1-32B-Instruct & \multirow{2}{*}{OLMo}    & \multirow{2}{*}{32B}               & \ding{55} \\
&& OLMo-3.1-32B-Think    &                          &                                     & \ding{51} \\
[1.5ex]
Specialized 
& 14B
& ChemDFM-v2.0-14B      &                          &                                     & \ding{55} \\
&& ChemDFM-R-14B         &                          &                                     & \ding{51} \\
& 24B
& Ether0                &                          &                                     & \ding{51} \\
[1.5ex]
Base models of  
& 14B
 & Qwen2.5-14B            & \multirow{5}{*}{Domain-specialized} & \multirow{3}{*}{14B}               & \ding{55} \\
~~~specialized
& 24B
 & Mistral-Small-24B     &                          & \multirow{2}{*}{24B}                & \ding{51} \\
[1.5ex]
Closed frontier 
& Unknown
& GPT-5.4-Mini               & \multirow{2}{*}{Frontier Closed}  & \multirow{2}{*}{---}               & \ding{51} \\
&& Claude-Haiku-4.5       &                          &                                     & \ding{51} \\
\bottomrule
\end{tabular}

\caption{All LLMs evaluated across five model families. The last column indicates if reasoning capability is present (\ding{51}) or absent (\ding{55}).
Qwen3 supports both modes whereas other models only support one. We also include the base model of the ChemDFMs (resp. Ether0), which is Qwen2.5-14B (resp. Mistral-Small-24B). Further details are provided in Section \ref{modelfamilies}.
}
\label{tab:models}
\end{table}

We evaluate the 16 models with 9 molecular representations across the 8 tasks. For each model-task configuration, we identify the representation with the highest mean score and then apply a paired bootstrap test \citep{koehn-2004-statistical} with 10,000 resamples between the per-instance scores of every other representation and the top scorer. 



\section{Results}
\label{results}

 We observe a consistent pattern: \textit{model performance is not representation-invariant}. The same molecule, expressed in different molecular representations, can lead to substantially different outcomes. This suggests that current LLMs' chemical competence is significantly influenced by the molecular text representation used. In addition, we notice that molecular text representations' embeddings from different LLMs across the nine representations show clear clustering amongst the different representations, indicating that the models fail to build representation-invariant internal encodings. We have provided the t-SNE plots of these embeddings in Appendix \ref{appendix_tsne}.


\subsection{Performance depends on representation}

\cut{
\begin{tcolorbox}[
    colback=warmpeach,
    colframe=warmborder,
    boxrule=0.5pt,
    arc=3pt,
    left=1pt, right=1pt, top=1pt, bottom=1pt
]
 No single molecular representation is uniformly optimal across tasks.
\end{tcolorbox}
}

\begin{table*}[htbp]
\centering
\scriptsize
\begin{tabular}{@{}l@{}cccccccccccc@{}}
\toprule
&&& \multicolumn{4}{c}{Property Estimation} \\\cmidrule(lr){4-7}
Representation & \tiny{Atom Count.} & \tiny{Func. Groups} & \tiny{logP} & \tiny{TPSA} & \tiny{HBD} & \tiny{HBA} & \tiny{Retrieval} & \tiny{Isomer Disc.} & \tiny{Cap.-to-Mol.} & \tiny{Tautomer} & \tiny{Protonation} & \tiny{\textbf{Total}} \\\midrule
Canonical SMILES & 0 & 3 & 7 & 7 & 3 & 5 & 13 & 0 & 9 & 7 & 12 & 66 \\
Isomeric SMILES & 0 & 2 & 10 & 7 & 2 & 6 & 11 & 0 & 8 & 5 & 12 & 63 \\
Randomized SMILES & 0 & 2 & 5 & 5 & 3 & 4 & 7 & 4 & 7 & 0 & 1 & 38 \\
DeepSMILES & 0 & 1 & 3 & 3 & 0 & 4 & 8 & 0 & 0 & 3 & 8 & 30 \\
IUPAC & 0 & 4 & 16 & 9 & 3 & 3 & 16 & 0 & 9 & 3 & 1 & 64 \\
SELFIES & 0 & 0 & 2 & 3 & 1 & 0 & 0 & 0 & 3 & 3 & 1 & 13 \\
CML & 9 & 9 & 2 & 6 & 13 & 8 & 1 & 14 & 0 & 13 & 6 & 81 \\
InChI & 12 & 0 & 9 & 8 & 1 & 15 & 9 & 1 & 0 & 9 & 8 & 72 \\
MolJSON & 9 & 4 & 3 & 7 & 0 & 13 & 3 & 13 & 4 & 12 & 7 & 75 \\
\bottomrule
\end{tabular}
\caption{Number of model configurations where each representation is among the best (paired bootstrap, 95\% confidence interval). Ties award a point to every indistinguishable representation.}
\label{repr_win_counts}

\end{table*}

Our central finding is that no single molecular representation is uniformly optimal across tasks. Table~\ref{repr_win_counts} summarizes the number of wins per representation across models for each task (detailed tables in Section~\ref{appendix_tables_results}). In aggregate, CML achieves the top score in 81 out of all 128 model-task configurations, more than any
other representation. MolJSON follows with 75, InChI with 72, and then canonical SMILES with 66.

This performance ranking is surprising. Representations predominant in LLM pretraining, such as the SMILES variants, are not the best. Instead CML and MolJSON are the top two. MolJSON is introduced recently (April 2026), so highly unlikely to be exposed to the LLMs we use (all released before April 2026). This contributes evidence that an LLM's performance on molecular tasks is not limited by its familiarity with a certain representation but by its ability to parse the implicit structure.

In atom counting (Table \ref{tab:atom_counting_accuracy}), the SMILES variants achieve high accuracies of 0.80 to 0.90.
Even so, InChI, CML and MolJSON surpass them: InChI wins with 12 LLMs, followed by CML and MolJSON at 9 each.  This could be due to the explicit mention of atom counts in InChI's formula layer (e.g., C9H8O4), whereas CML and MolJSON only mention each atom explicitly. SELFIES is the worst, losing with all 16 LLMs.

 Functional-group identification is led by CML, followed by both IUPAC and MolJSON in 4 models (Tables \ref{repr_win_counts} and \ref{tab:functional_groups_macro_f1}). Two patterns arise here: CML supports graph reasoning, which is helpful in finding functional groups that are substructures in the molecular graph, while IUPAC exposes the functional groups directly through its functional group morphemes (e.g. carboxylic acid, ester, amide). 
 
 
In property estimation (Tables \ref{property_estimation_logp_spearman_rho} to \ref{tab:property_estimation_hba_exact_match}), SMILES variants and IUPAC dominate in LogP and TPSA regression while being average in hydrogen bond donor (HBD) and acceptor (HBA) counts. For HBA and HBD counts, structured representations dominate.
With Qwen3-4B, CML is twice as good as SMILES, increasing accuracy to 0.79 from 0.30;  with GPT-5.4-Mini, to 0.98 from 0.81. MolJSON, while generally trailing CML, is better than the other representations in the HBA and HBD counts.
The results reinforce the intuition that the structured text representations are apt at tracking the various atoms and bonds while SMILES and IUPAC are apt for the regression tasks.

Structured text representations, CML and MolJSON, dominate isomer discrimination and tautomer recognition (Tables \ref{repr_win_counts}, \ref{tab:isomer_discrimination_accuracy} and \ref{tab:tautomer_recognition_accuracy}). These tasks require graph understanding and fine-grained structural disambiguation, which are both facilitated by CML and MolJSON. Canonical and isomeric SMILES are the top two representations for protonation-state recognition: SMILES variants present charges as visible characters, for example \texttt{CC(=O)O} (acid) versus \texttt{CC(=O)[O-]} (anion). IUPAC and SELFIES are the worst and collapse to random performance with many LLMs. 
These two do not explicitly mention the charge: IUPAC names are neutral by definition, and SELFIES  grammar enforces valence constraints by construction.

However, IUPAC shines in molecule retrieval (Table \ref{tab:molecule_retrieval_accuracy}): it is the best representation for all the 16 LLMs. Canonical SMILES and isomeric SMILES trail closely behind, with 13 and 11 LLMs, respectively. IUPAC's dominant performance highlights its morphemic structure, which is directly related to the natural language descriptions of molecules present in molecule retrieval questions. Also, IUPAC performs the best in caption-to-molecule generation alongside the SMILES variants (see also Section~\ref{cap2mol}).

Therefore, the optimal representation depends on the task. Counting and structural-disambiguation tasks benefit from explicit, addressable formats (CML, MolJSON, InChI), while description-matching tasks favour IUPAC.

\subsection{Chemistry-Specialized Post-Training limits robustness across representations}
\cut{
\begin{tcolorbox}[
    colback=warmpeach,
    colframe=warmborder,
    boxrule=0.5pt,
    arc=3pt,
    left=1pt, right=1pt, top=1pt, bottom=1pt
]
Chemistry-specialized models do not necessarily boost performance compared to their base models, which surfaces the need for reporting performance across multiple representations.
\end{tcolorbox}
}

To understand the capabilities of chemistry-specialized models and isolate the effect of domain post-training from the base model's general capability, we compare the specialized models with their respective base models. 
We find that chemistry-specialized models do not necessarily perform better than their base models across representations, which surfaces the need for reporting performance across multiple representations.

ChemDFM-R-14B with canonical SMILES is the best in the entire study for caption-to-molecule generation in terms of exact match 
(Table~\ref{tab:caption_to_molecule_exact_match}), validity rate (Table \ref{tab:caption_to_molecule_validity_rate}), 
and  FCD (Table~\ref{tab:caption_to_molecule_fcd}). On SMILES representations, the ChemDFMs improve across tasks over its base model, Qwen2.5-14B: on molecule retrieval, accuracy rises from 0.27 to 0.72 (resp.\ 0.80) for ChemDFM-v2.0 (resp.\ ChemDFM-R); on tautomer recognition, from 0.26 to 0.88 (resp.\ 0.92); and on protonation state recognition, from 0.28 to 0.88 (resp.\ 0.95).

However, these improvements are not uniform across representations. The ChemDFMs score near zero with CML and InChI for molecule generation. This suggests that the fine-tuning data, likely dominated by SMILES, strengthens SMILES-based chemical reasoning at the cost of generalization to markup and line-notation formats. Compared to Qwen2.5-14B in property estimation, ChemDFM-v2.0 is worse with MolJSON, 
CML and especially InChI.
In atom counting,  ChemDFM-v2.0 is worse with IUPAC, 
SELFIES, 
and especially InChI. 

Hence, fine-tuning Qwen2.5-14B seems to erode structured-format understanding for a boost in SMILES fluency, and the poor performance is most visible on tasks where the model needs to parse the representations rather than just pattern-match. This may be attributed to the \textit{catastrophic forgetting} when LLMs are fine-tuned or post-trained on tasks with a shift in distribution, making them unable to retain past knowledge or capability \citep{Kirkpatrick_2017, liao-etal-2025-exploring}.  

Similarly, Ether0-24B performs worse than its base model (Mistral-Small-24B) on essentially every task in our benchmark. 
Details in Section \ref{ether0_app}. 

Therefore, evaluating chemistry LLMs on SMILES-only tasks systematically rewards specialization that does not generalize. Reporting performance across multiple representations, as we do here, is necessary to surface trade-offs.
 
\subsection{Caption-to-molecule differentiates representations best}
\label{cap2mol}
Caption-to-molecule generation
(Tables~\ref{tab:caption_to_molecule_validity_rate}--\ref{tab:caption_to_molecule_fcd}) requires constructing an
exact molecular graph from a natural language description. Under the paired bootstrap analysis (Table \ref{repr_win_counts}), canonical SMILES and IUPAC jointly lead (9 points), followed by isomeric SMILES (8) and randomized SMILES (7). Three key trends emerge from this task.

First, validity does not imply structural correctness. ChemDFM-R-14B produces 97.6\% valid molecules with canonical SMILES, yet its Morgan Tanimoto similarity is 0.68 (Table \ref{tab:caption_to_molecule_mean_tanimoto}). Similarly, GPT-5.4-Mini produces 91.6\% valid molecules with canonical SMILES, yet its Morgan Tanimoto similarity with canonical SMILES is 0.61. 

Next, robustness across representations distinguishes frontier from specialized models.  ChemDFM-R-14B's good results with canonical SMILES come at the cost of complete failure with CML, MolJSON and InChI, where it produces no valid generations. GPT-5.4-Mini's canonical SMILES scores are slightly lower compared to ChemDFM-R-14B but the model produces nontrivial valid generations for structured text representations (Morgan Tanimoto similarity for CML is 0.24; for MolJSON, 0.30). Hence, frontier models are more robust in generation, while ChemDFM-R's good results are limited to SMILES.

Third, IUPAC's natural language and morphemic structure is particularly relevant to molecule generation. As mentioned previously, the ChemDFMs achieve the best across the generation metrics when using canonical SMILES. However, ChemDFMs' performance with other representations drops severely. This could be attributed to ChemDFMs being post-trained heavily on chemical text that contains a large amount of SMILES strings. Meanwhile, across different generation metrics and non-ChemDFM models, IUPAC gives the best performance. This is due to ChEBI-20's captions referencing systematic-name fragments that are directly relevant to the IUPAC nomenclature.

\subsection{LLM-as-a-judge reveals major sources of error in generation across representations}
\label{llmjudge}

\begin{figure}
  \centering
  \includegraphics[width=\columnwidth]{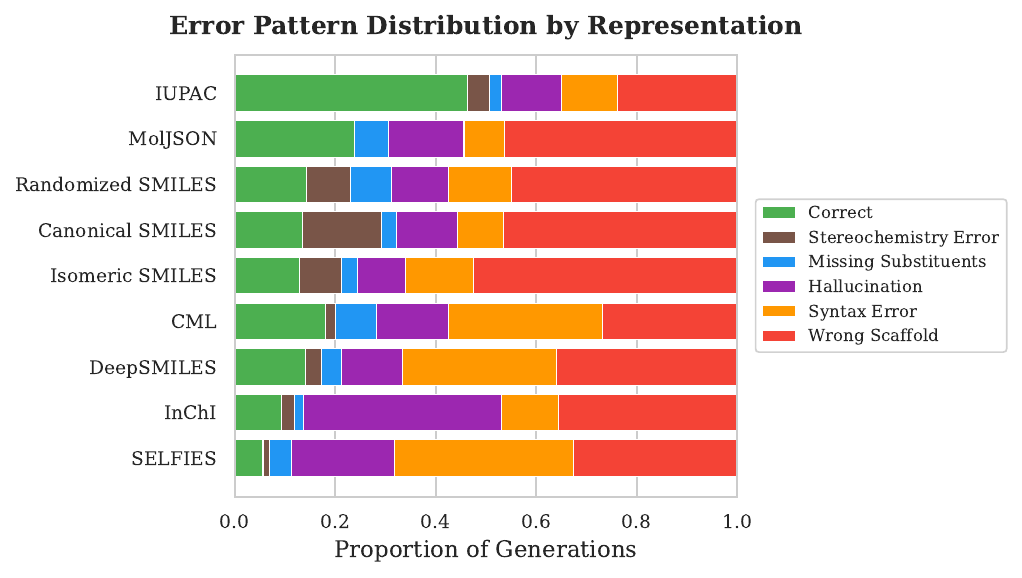}
  \caption{
    Proportion of error modes in molecule generation across various representations as determined by Gemini 3 Flash.
  }
  \label{qual_b6_error_patterns}
\end{figure}

\begin{figure}
  \centering
  \includegraphics[width=\columnwidth]{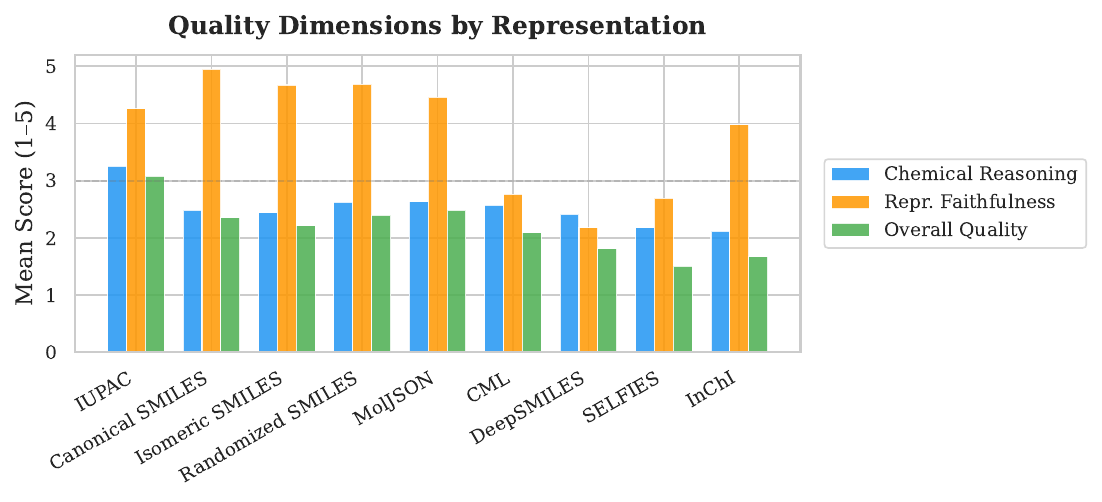}
  \caption{
    Gemini 3 Flash's scoring (from 1 to 5) across representations in the molecule generation task, based on faithfulness, chemical reasoning, and overall quality.
  }
  \label{qual_b6_dimensions}
\end{figure}

While validity rate, exact match, and the Tanimoto similarity metrics give an understanding of the quality and correctness of the generations, we also seek to know exactly where each model goes wrong with different representations. We use Gemini 3 Flash to judge the generations for 10 captions from ChEBI-20, for all models and representations. The LLM-as-a-judge categorizes errors into five types: stereochemistry, missing substituents, wrong scaffold, hallucination and/or syntax error (Figure \ref{qual_b6_error_patterns}). It scores the generations against representation faithfulness, chemical reasoning, and overall quality, each on a 1--to--5 scale (Figure \ref{qual_b6_dimensions}). Despite the small sample size, we can glean some insights into the errors for each representation.

IUPAC leads the performance with 50\% correct generations, while SELFIES and DeepSMILES are the worst. Canonical SMILES has the highest representation faithfulness but relatively low overall quality; it has a significant issue with wrong scaffolds and chemical reasoning, despite SMILES being most represented in chemical training data. When prompted to generate in CML, DeepSMILES or SELFIES, LLMs score low on representation faithfulness, with syntax being the major error. IUPAC is the only representation where both reasoning and faithfulness scores are high, so we should prioritise IUPAC names for molecule generation. MolJSON has a higher representation faithfulness score while it closely trails IUPAC in terms of chemical reasoning and overall quality.

While the hallucination errors are less frequent compared to other error types, they remain a major issue with current LLMs \citep{hallu-rawte-etal-2023-troubling}. For instance, ChemDFM-R-14B just generates a string of multiple `C's, and OLMo-3.1-32B-Think, when asked for a CML output, outputs a malformed SMILES string. On the other hand, Ether0 wrongly identifies pneumocandin, an antifungal drug \citep{pneumocandin}, as a chemical weapon and refuses to proceed with the response. Given that hallucination errors are diverse, using LLM-as-a-judge in addition to rigid metrics helps to analyse the qualitative nature of these errors. 

\begin{figure}
  \centering
  \includegraphics[width=\columnwidth]{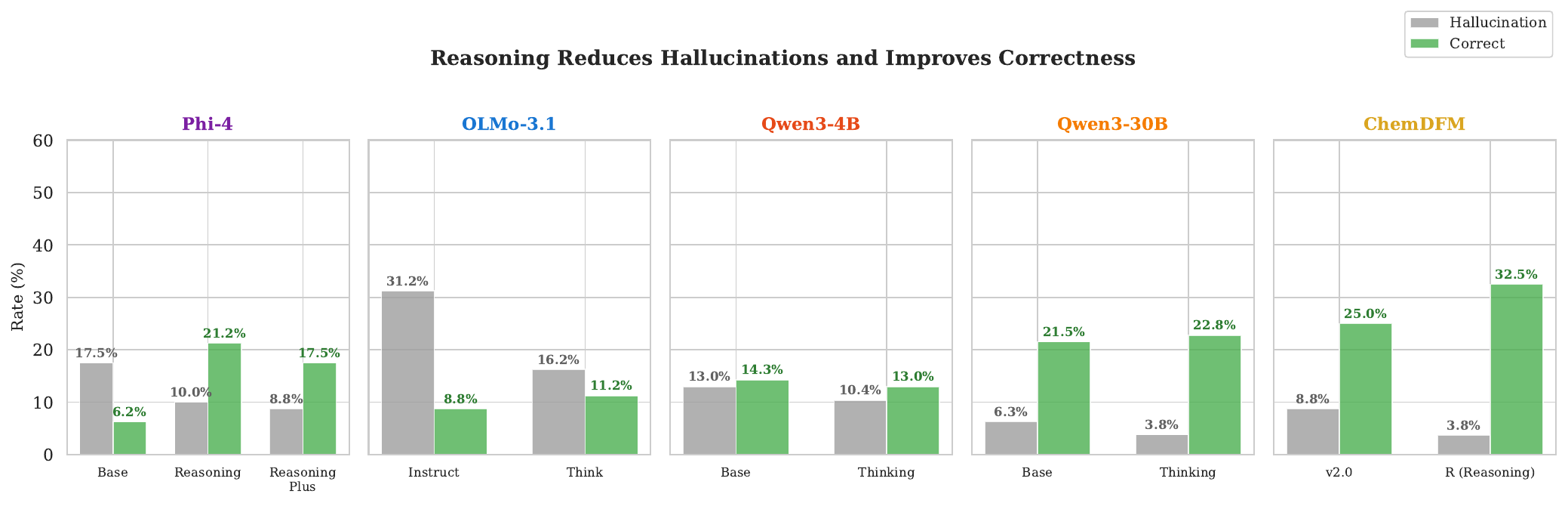}
  \caption{
    Reasoning models reduce hallucination.
  }
  \label{qual_b6_hallucination_drop}
\end{figure}

Finally, reasoning LLMs reduce hallucination (Figure \ref{qual_b6_hallucination_drop}). For OLMo, the reasoning variant halves the hallucination rate to 16.2\% from the non-reasoning variant's 31.2\%; and for ChemDFM, the reasoning variant halves the hallucination rate to 3.8\% from the non-reasoning variant's 8.8\%. Moreover, reasoning LLMs also lead to more correct responses generally. The observations carry over to Phi-4 and Qwen3 models, though the improvements are not as pronounced.

\section{Mechanistic Analysis}
\label{sec:mechanistic}

To explain the representation-dependence in performance in Section \ref{results}, we conduct mechanistic analysis using the Qwen3-4B (36 transformer layers and 151,669 token vocabulary). We use this model as it has open weights, it is small, and all its parameters are active. 
We conduct (1) a tokenization audit that examines how each representation is segmented, (2) linear probing of hidden states to measure what chemical information the model encodes internally, and (3) attention analysis to locate where the model attends for each representation.


\subsection{Tokenization Audit}
\label{sec:s1_tokenization}

We tokenize 1,000 molecules from ChEBI-20 in all nine representations and measure the total token count of the actual molecular representation and the length of reasoning tokens. 

Representations vary by an order of magnitude in token cost. DeepSMILES is the most compact (median 25 tokens), followed by canonical SMILES (28), randomized SMILES (30), and isomeric SMILES (36). IUPAC names require 47 tokens on average. SELFIES is 3.6 times longer than canonical SMILES (median 101). InChI, MolJSON and CML are the longest. Such representation lengths are costly given fixed context windows and quadratic attention scaling. However, the reasoning token length presents a trade-off (Figure \ref{reasoning_len__atom_counting}): explicit, verbose representations like CML and MolJSON
result in shorter reasoning token length. 

\begin{figure}
  \centering
  \includegraphics[width=\columnwidth]{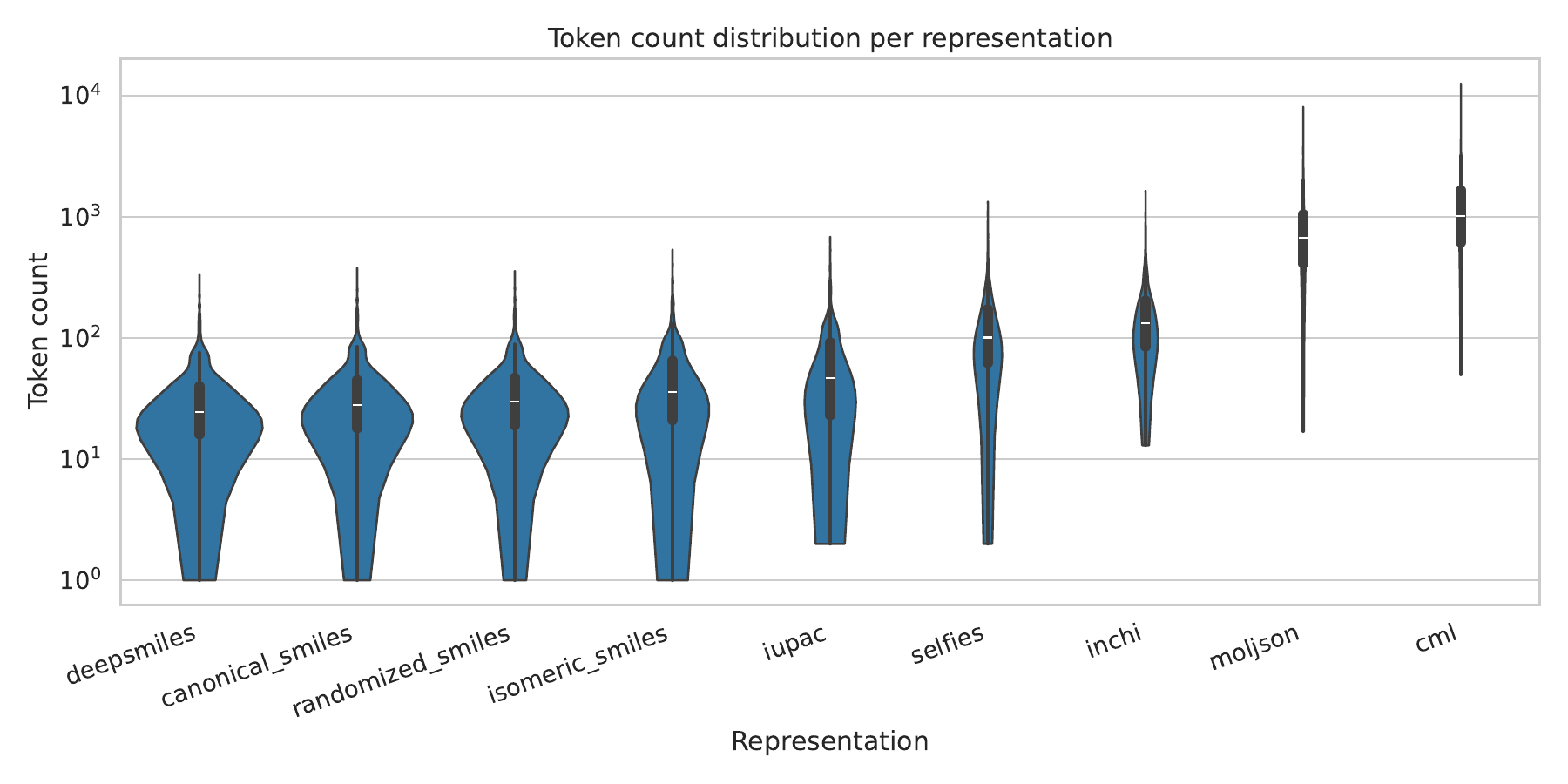}
  \caption{
    Token count distribution across 9 representations using Qwen3's tokenizer.
  }
  \label{token_count_violin}
\end{figure}

\subsection{Linear Probing of Hidden States}
\label{sec:s2_probing}

\begin{figure}
  \centering
  \includegraphics[width=\columnwidth]{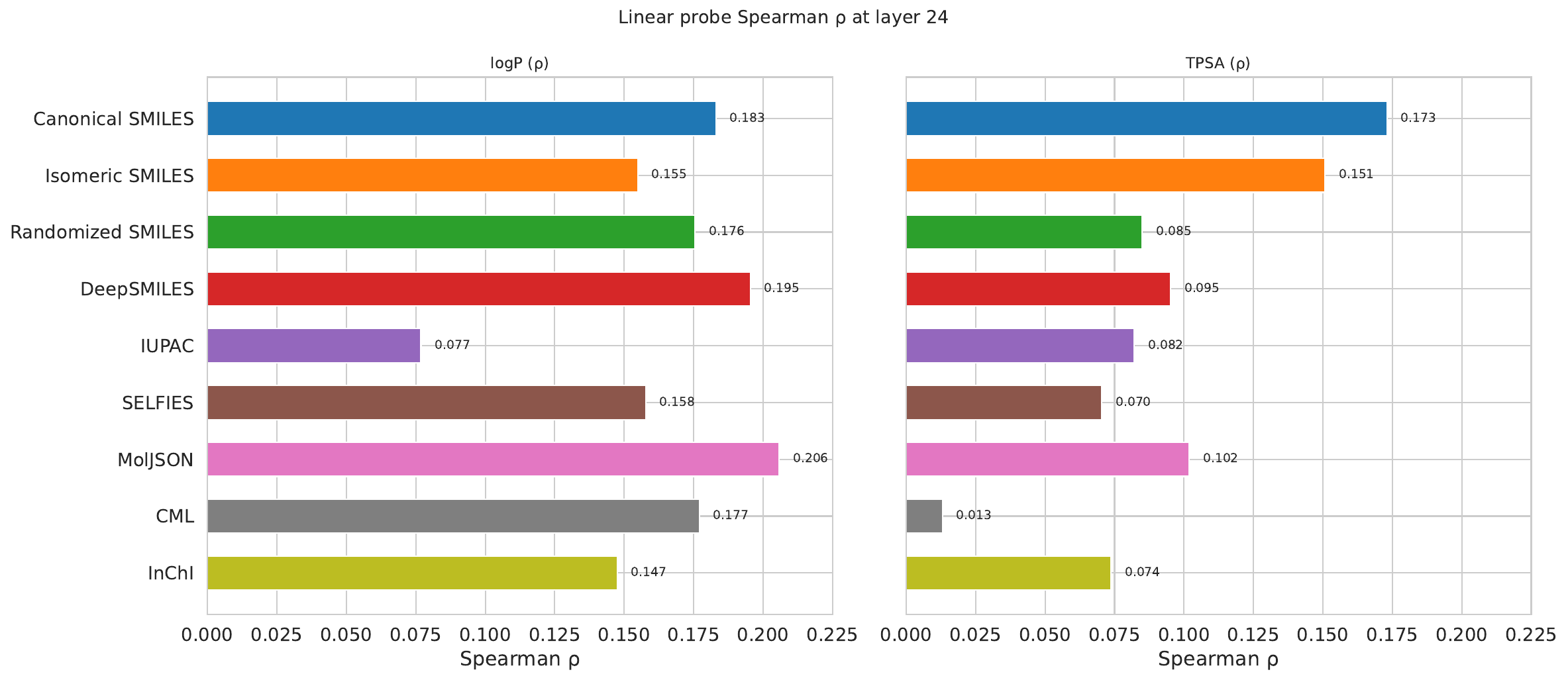}
  \caption{
   Linear probing performance of Qwen3-4B across all representations for molecular weight, log P, and TPSA prediction
  }
  \label{linear_probe_bar}
\end{figure}

Using the same 1,000 molecules from Section \ref{sec:s1_tokenization}, 
we use linear probes to see if the hidden embeddings can already estimate molecular properties.
This can elicit if the representation already provides sufficient signals in the intermediate layers before the final predictions.
We train linear probes (ridge regression for continuous targets) on the hidden states of each layer to estimate three molecular properties: molecular weight (MW), LogP and TPSA. We use the intermediate embeddings (the 24th layer) as inputs to train the probes.

The SMILES variants have some of the best performances.
This is expected, given that they are likely present during pre-training. 
However, MolJSON, which is created after Qwen3-4B was released, results in the best performances. CML, though an explicit graph-based representation like MolJSON, underperforms to varying degrees on all probing tasks, indicating that the internal representations from the XML format are not sufficiently informative. However, the performance of Qwen3-4B when prompted on logP and TPSA estimation leads to much higher Spearman correlation across all representations (Tables \ref{tab:property_estimation_tpsa_spearman_rho} and \ref{property_estimation_logp_spearman_rho}).
This disconnect between benchmark performance and linear probe performance suggests two points:
(1) LLMs can better estimate molecular properties through the generation process during inference instead of storing them as linearly accessible features in its hidden states \citep{belinkov, gao-etal-2024-insights}, and (2) the representation choice affects the LLM's ability to leverage encoded knowledge during generation as opposed to just impacting the richness of the encodings.

\subsection{Attention analysis}
\label{appendix_attention}

We extract attention weights from all 36 layers and 32 heads for 50 selected molecules across all representations. We measure two quantities: (i) \emph{last token to molecule attention} (Figure \ref{last-tok-mol}), which is the fraction of the final prediction token's attention towards the molecule tokens, and (ii) \emph{within-molecule attention} (Figure \ref{within-mol}), which is the self-attention amongst the molecule's tokens. We perform this analysis from layers 17 to 24. We chose these layers as they form the intermediate layers and exhibit reasonable performance during linear probing.

CML and MolJSON have the largest last token to molecule attention, nearly 5 to 10 times that of canonical SMILES. This is consistent with the long length of the explicit graph representations: the model must look further back to gather molecular information. 
By contrast, SMILES variants have 20 times larger within-molecule attention compared to MolJSON and CML, suggesting that compact representations encourage richer token-to-token interaction within the molecular string.



\section{Conclusion}

We introduced MolRepBench, a systematic evaluation of nine molecular text representations across eight tasks and 16 LLMs spanning open-weight, reasoning-tuned, domain-specialized, and closed frontier models.
Our results show that \textbf{LLM-based chemical reasoning is not representation-invariant}. The same molecule, expressed in different molecular representations, can yield significantly different outcomes. CML and MolJSON dominate tasks that require understanding of molecular graphs: atom counting, isomer discrimination, and tautomer recognition. IUPAC names dominate semantically grounded tasks such as molecule retrieval and caption-to-molecule generation, where their morphemic structure highlights key chemical units. 
Despite SMILES prevalence in pretraining corpora, they are rarely optimal. 


Our findings suggest to adopt the following:
(1) benchmarks should report performances across multiple molecular representations, 
(2) an LLM-judge should be used to qualitatively analyse the errors (in LLM outputs) beyond quantitative metrics such as validity rate and Tanimoto similarities, and
(3) molecular applications should select the molecular representation based on the task.

\section*{Limitations}


Our goal is to understand how LLMs use the various text representations to solve various tasks, and we have strived to cover a wide number of chemistry tasks and representations relevant to current cheminformatics tools and LLMs. 
While more chemistry tasks can be added to enhance coverage, our benchmark have already led to clear and actionable insights that could enable robust use of LLMs for chemistry.
The study have solely focused on the use of text representations with LLMs and have not explored multi-modal models, which are becoming more common.
Future work can understand how multi-modal models perform when given 1D, 2D and 3D representations of molecules.

\section*{Ethics Statement}
Our work does not involve any human subjects. We use publicly available datasets, open-source LLMs, and openly accessible frontier LLMs. We intend to understand the effect of molecular text representations with LLMs, and there is potential for dual use. Our findings could aid medicinal chemists working on drug discovery or nefarious users trying to create harmful molecules.

\section*{Acknowledgements}
A.R.’s PhD program is supported by the Agency for Science, Technology, and Research and the SABS
R3 CDT program via the Engineering and Physical Sciences Research Council. This work was done during A.R.'s internship at DSO National Laboratories as part of the DSO-AISG Incentive Award. We would like to thank DSO and AI Singapore for the computational resources, which played a significant role in this research. We would also like to thank Dr Hongtao Zhao, Dr Christian Tyrchan, Dr Eva Nittinger, Prof. Charlotte M. Deane, and Prof. Michael M. Bronstein for their advice. 

\bibliographystyle{acl_natbib}
\bibliography{emnlp2023}

@inproceedings{christofidellis2023unifying,
  title = 	 {Unifying Molecular and Textual Representations via Multi-task Language Modelling},
  author =       {Christofidellis, Dimitrios and Giannone, Giorgio and Born, Jannis and Winther, Ole and Laino, Teodoro and Manica, Matteo},
  booktitle = 	 {Proceedings of the 40th International Conference on Machine Learning},
  pages = 	 {6140--6157},
  year = 	 {2023},
  volume = 	 {202},
  series = 	 {Proceedings of Machine Learning Research},
  publisher =    {PMLR},
  url = 	 {https://proceedings.mlr.press/v202/christofidellis23a.html},
}

@article{Jablonka2024LeveragingLL,
  title={Leveraging large language models for predictive chemistry},
  author={Kevin Maik Jablonka and Philippe Schwaller and Andres Ortega‐Guerrero and Berend Smit},
  journal={Nature Machine Intelligence},
  year={2024},
  volume={6},
  pages={161 - 169},
  url={https://api.semanticscholar.org/CorpusID:267538205}
}

@article{Li_2024,
   title={Empowering Molecule Discovery for Molecule-Caption Translation With Large Language Models: A ChatGPT Perspective},
   volume={36},
   ISSN={2326-3865},
   url={http://dx.doi.org/10.1109/TKDE.2024.3393356},
   DOI={10.1109/tkde.2024.3393356},
   number={11},
   journal={IEEE Transactions on Knowledge and Data Engineering},
   publisher={Institute of Electrical and Electronics Engineers (IEEE)},
   author={Li, Jiatong and Liu, Yunqing and Fan, Wenqi and Wei, Xiao-Yong and Liu, Hui and Tang, Jiliang and Li, Qing},
   year={2024},
   month=Nov, pages={6071–6083} }

@inproceedings{guo_bench,
 author = {Guo, Taicheng and Guo, kehan and Nan, Bozhao and Liang, Zhenwen and Guo, Zhichun and Chawla, Nitesh and Wiest, Olaf and Zhang, Xiangliang},
 booktitle = {Advances in Neural Information Processing Systems},
 editor = {A. Oh and T. Naumann and A. Globerson and K. Saenko and M. Hardt and S. Levine},
 pages = {59662--59688},
 publisher = {Curran Associates, Inc.},
 title = {What can Large Language Models do in chemistry? A comprehensive benchmark on eight tasks},
 url = {https://proceedings.neurips.cc/paper_files/paper/2023/file/bbb330189ce02be00cf7346167028ab1-Paper-Datasets_and_Benchmarks.pdf},
 volume = {36},
 year = {2023}
}

@article{nach0,
   title={nach0: multimodal natural and chemical languages foundation model},
   volume={15},
   ISSN={2041-6539},
   url={http://dx.doi.org/10.1039/D4SC00966E},
   DOI={10.1039/d4sc00966e},
   number={22},
   journal={Chemical Science},
   publisher={Royal Society of Chemistry (RSC)},
   author={Livne, Micha and Miftahutdinov, Zulfat and Tutubalina, Elena and Kuznetsov, Maksim and Polykovskiy, Daniil and Brundyn, Annika and Jhunjhunwala, Aastha and Costa, Anthony and Aliper, Alex and Aspuru-Guzik, Alán and Zhavoronkov, Alex},
   year={2024},
   pages={8380–8389} }

@misc{taylor2022galacticalargelanguagemodel,
      title={Galactica: A Large Language Model for Science}, 
      author={Ross Taylor and Marcin Kardas and Guillem Cucurull and Thomas Scialom and Anthony Hartshorn and Elvis Saravia and Andrew Poulton and Viktor Kerkez and Robert Stojnic},
      year={2022},
      eprint={2211.09085},
      archivePrefix={arXiv},
      primaryClass={cs.CL},
      url={https://arxiv.org/abs/2211.09085}, 
}

@inproceedings{hallu-rawte-etal-2023-troubling,
    title = "The Troubling Emergence of Hallucination in Large Language Models - An Extensive Definition, Quantification, and Prescriptive Remediations",
    author = "Rawte, Vipula  and
      Chakraborty, Swagata  and
      Pathak, Agnibh  and
      Sarkar, Anubhav  and
      Tonmoy, S.M Towhidul Islam  and
      Chadha, Aman  and
      Sheth, Amit  and
      Das, Amitava",
    editor = "Bouamor, Houda  and
      Pino, Juan  and
      Bali, Kalika",
    booktitle = "Proceedings of the 2023 Conference on Empirical Methods in Natural Language Processing",
    month = dec,
    year = "2023",
    address = "Singapore",
    publisher = "Association for Computational Linguistics",
    url = "https://aclanthology.org/2023.emnlp-main.155/",
    doi = "10.18653/v1/2023.emnlp-main.155",
    pages = "2541--2573"
}

@inproceedings{edwards-etal-2022-translation,
    title = "Translation between Molecules and Natural Language",
    author = "Edwards, Carl  and
      Lai, Tuan  and
      Ros, Kevin  and
      Honke, Garrett  and
      Cho, Kyunghyun  and
      Ji, Heng",
    editor = "Goldberg, Yoav  and
      Kozareva, Zornitsa  and
      Zhang, Yue",
    booktitle = "Proceedings of the 2022 Conference on Empirical Methods in Natural Language Processing",
    month = dec,
    year = "2022",
    address = "Abu Dhabi, United Arab Emirates",
    publisher = "Association for Computational Linguistics",
    url = "https://aclanthology.org/2022.emnlp-main.26/",
    doi = "10.18653/v1/2022.emnlp-main.26",
    pages = "375--413"
}

@article{weininger1988smiles,
  title={SMILES, a chemical language and information system. 1. Introduction to methodology and encoding rules},
  author={David Weininger},
  journal={J. Chem. Inf. Comput. Sci.},
  year={1988},
  volume={28},
  pages={31-36},
  url={https://api.semanticscholar.org/CorpusID:5445756}
}

@article{wiley_rev2,
author = {Raghunathan, Shampa and Priyakumar, U. Deva},
title = {Molecular representations for machine learning applications in chemistry},
journal = {International Journal of Quantum Chemistry},
volume = {122},
number = {7},
pages = {e26870},
doi = {https://doi.org/10.1002/qua.26870},
url = {https://onlinelibrary.wiley.com/doi/abs/10.1002/qua.26870},
eprint = {https://onlinelibrary.wiley.com/doi/pdf/10.1002/qua.26870},
year = {2022}
}

@article{az_rev,
author = {David, Laurianne and Thakkar, Amol and Mercado, Rocío and Engkvist, Ola},
year = {2020},
month = {09},
pages = {},
title = {Molecular representations in AI-driven drug discovery: a review and practical guide},
volume = {12},
journal = {Journal of Cheminformatics},
doi = {10.1186/s13321-020-00460-5}
}

@article{wiley_rev,
author = {Wigh, Daniel S. and Goodman, Jonathan M. and Lapkin, Alexei A.},
title = {A review of molecular representation in the age of machine learning},
journal = {WIREs Computational Molecular Science},
volume = {12},
number = {5},
pages = {e1603},
doi = {https://doi.org/10.1002/wcms.1603},
url = {https://wires.onlinelibrary.wiley.com/doi/abs/10.1002/wcms.1603},
eprint = {https://wires.onlinelibrary.wiley.com/doi/pdf/10.1002/wcms.1603},
year = {2022}
}

@article{krenn2020selfies,
   title={Self-referencing embedded strings (SELFIES): A 100\% robust molecular string representation},
   volume={1},
   ISSN={2632-2153},
   url={http://dx.doi.org/10.1088/2632-2153/aba947},
   DOI={10.1088/2632-2153/aba947},
   number={4},
   journal={Machine Learning: Science and Technology},
   publisher={IOP Publishing},
   author={Krenn, Mario and Häse, Florian and Nigam, AkshatKumar and Friederich, Pascal and Aspuru-Guzik, Alan},
   year={2020},
   month=Oct,
   pages={045024}
}

@article{inchi_mcnaught2006iupac,
  title={The IUPAC international chemical identifier: InChl-A new standard for molecular informatics},
  author={Alan McNaught},
  journal={Chemistry international},
  year={2006},
  volume={28},
  pages={12-14},
  url={https://api.semanticscholar.org/CorpusID:89057853}
}

@article{chemdoodle,
author = {Burger, Melanie},
year = {2015},
month = {07},
pages = {35},
title = {ChemDoodle Web Components: HTML5 toolkit for chemical graphics, interfaces, and informatics},
volume = {7},
journal = {Journal of cheminformatics},
doi = {10.1186/s13321-015-0085-3}
}

@misc{blinding,
      title={In-Context Molecular Property Prediction with LLMs: A Blinding Study on Memorization and Knowledge Conflicts}, 
      author={Matthias Busch and Marius Tacke and Sviatlana V. Lamaka and Mikhail L. Zheludkevich and Christian J. Cyron and Christian Feiler and Roland C. Aydin},
      year={2026},
      eprint={2603.25857},
      archivePrefix={arXiv},
      primaryClass={cs.LG},
      url={https://arxiv.org/abs/2603.25857}, 
}

@misc{cheng2025surveydatacontaminationlarge,
      title={A Survey on Data Contamination for Large Language Models}, 
      author={Yuxing Cheng and Yi Chang and Yuan Wu},
      year={2025},
      eprint={2502.14425},
      archivePrefix={arXiv},
      primaryClass={cs.CL},
      url={https://arxiv.org/abs/2502.14425}, 
}

@inproceedings{sainz-etal-2023-nlp,
    title = "{NLP} Evaluation in trouble: On the Need to Measure {LLM} Data Contamination for each Benchmark",
    author = "Sainz, Oscar  and
      Campos, Jon  and
      Garc{\'i}a-Ferrero, Iker  and
      Etxaniz, Julen  and
      de Lacalle, Oier Lopez  and
      Agirre, Eneko",
    editor = "Bouamor, Houda  and
      Pino, Juan  and
      Bali, Kalika",
    booktitle = "Findings of the Association for Computational Linguistics: EMNLP 2023",
    month = dec,
    year = "2023",
    address = "Singapore",
    publisher = "Association for Computational Linguistics",
    url = "https://aclanthology.org/2023.findings-emnlp.722/",
    doi = "10.18653/v1/2023.findings-emnlp.722",
    pages = "10776--10787"
}

@inproceedings{Baker2025MolecularSR,
  title={Molecular String Representation Preferences in Pretrained LLMs: A Comparative Study in Zero- \& Few-Shot Molecular Property Prediction},
  author={George Arthur Baker and Mario Sanz-Guerrero and Katharina von der Wense},
  booktitle={Conference on Empirical Methods in Natural Language Processing},
  year={2025},
  url={https://api.semanticscholar.org/CorpusID:282892919}
}

@article{Zhong2024BenchmarkingLL,
  title={Benchmarking Large Language Models for Molecule Prediction Tasks},
  author={Zhiqiang Zhong and Kuangyu Zhou and Davide Mottin},
  journal={ArXiv},
  year={2024},
  volume={abs/2403.05075},
  url={https://api.semanticscholar.org/CorpusID:268297055}
}

@article{Gupta_2025,
   title={Benchmarking Large Language Models for Polymer Property Predictions},
   ISSN={1521-3927},
   url={http://dx.doi.org/10.1002/marc.202500388},
   DOI={10.1002/marc.202500388},
   journal={Macromolecular Rapid Communications},
   publisher={Wiley},
   author={Gupta, Sonakshi and Mahmood, Akhlak and Shukla, Shivank and Ramprasad, Rampi},
   year={2025},
   month=Oct }

@inproceedings{yauney-mimno-2021-comparing,
    title = "Comparing Text Representations: {A} Theory-Driven Approach",
    author = "Yauney, Gregory  and
      Mimno, David",
    editor = "Moens, Marie-Francine  and
      Huang, Xuanjing  and
      Specia, Lucia  and
      Yih, Scott Wen-tau",
    booktitle = "Proceedings of the 2021 Conference on Empirical Methods in Natural Language Processing",
    month = nov,
    year = "2021",
    address = "Online and Punta Cana, Dominican Republic",
    publisher = "Association for Computational Linguistics",
    url = "https://aclanthology.org/2021.emnlp-main.449/",
    doi = "10.18653/v1/2021.emnlp-main.449",
    pages = "5527--5539"
}

@misc{xu2025reimaginesymbolicbenchmarksynthesis,
      title={RE-IMAGINE: Symbolic Benchmark Synthesis for Reasoning Evaluation}, 
      author={Xinnuo Xu and Rachel Lawrence and Kshitij Dubey and Atharva Pandey and Risa Ueno and Fabian Falck and Aditya V. Nori and Rahul Sharma and Amit Sharma and Javier Gonzalez},
      year={2025},
      eprint={2506.15455},
      archivePrefix={arXiv},
      primaryClass={cs.CL},
      url={https://arxiv.org/abs/2506.15455}, 
}

@misc{moleculenet,
      title={MoleculeNet: A Benchmark for Molecular Machine Learning}, 
      author={Zhenqin Wu and Bharath Ramsundar and Evan N. Feinberg and Joseph Gomes and Caleb Geniesse and Aneesh S. Pappu and Karl Leswing and Vijay Pande},
      year={2018},
      eprint={1703.00564},
      archivePrefix={arXiv},
      primaryClass={cs.LG},
      url={https://arxiv.org/abs/1703.00564}, 
}

@inproceedings{al-shaibani-ahmad-2023-consonant,
    title = "Consonant is all you need: a compact representation of {E}nglish text for efficient {NLP}",
    author = "Al-shaibani, Maged S.  and
      Ahmad, Irfan",
    editor = "Bouamor, Houda  and
      Pino, Juan  and
      Bali, Kalika",
    booktitle = "Findings of the Association for Computational Linguistics: EMNLP 2023",
    month = dec,
    year = "2023",
    address = "Singapore",
    publisher = "Association for Computational Linguistics",
    url = "https://aclanthology.org/2023.findings-emnlp.775/",
    doi = "10.18653/v1/2023.findings-emnlp.775",
    pages = "11578--11588"
}

@misc{jacobs2026regressionlargelanguagemodels,
      title={Regression with Large Language Models for Materials and Molecular Property Prediction}, 
      author={Ryan Jacobs and Maciej P. Polak and Lane E. Schultz and Hamed Mahdavi and Vasant Honavar and Dane Morgan},
      year={2026},
      eprint={2409.06080},
      archivePrefix={arXiv},
      primaryClass={cond-mat.mtrl-sci},
      url={https://arxiv.org/abs/2409.06080}, 
}

@article{o2018deepsmiles,
  title={DeepSMILES: An Adaptation of SMILES for Use in Machine-Learning of Chemical Structures},
  author={Noel M. O'Boyle and Andrew Dalke},
  journal={ChemRxiv},
  year={2018},
  url={https://api.semanticscholar.org/CorpusID:69388493}
}

@misc{mirza2024large,
      title={Are large language models superhuman chemists?}, 
      author={Adrian Mirza and Nawaf Alampara and Sreekanth Kunchapu and Martiño Ríos-García and Benedict Emoekabu and Aswanth Krishnan and Tanya Gupta and Mara Schilling-Wilhelmi and Macjonathan Okereke and Anagha Aneesh and Amir Mohammad Elahi and Mehrdad Asgari and Juliane Eberhardt and Hani M. Elbeheiry and María Victoria Gil and Maximilian Greiner and Caroline T. Holick and Christina Glaubitz and Tim Hoffmann and Abdelrahman Ibrahim and Lea C. Klepsch and Yannik Köster and Fabian Alexander Kreth and Jakob Meyer and Santiago Miret and Jan Matthias Peschel and Michael Ringleb and Nicole Roesner and Johanna Schreiber and Ulrich S. Schubert and Leanne M. Stafast and Dinga Wonanke and Michael Pieler and Philippe Schwaller and Kevin Maik Jablonka},
      year={2024},
      eprint={2404.01475},
      archivePrefix={arXiv},
      primaryClass={cs.LG},
      url={https://arxiv.org/abs/2404.01475}, 
}

@article{liu2023moleculestm,
    title = "Multi-modal Molecule Structure-text Model for Text-based Retrieval and Editing",
    author = "Liu, Shengchao and Nie, Weili and Wang, Chengpeng and Lu, Jiarui and Qiao, Zhuoran and Liu, Ling and Tang, Jian and Xiao, Chaowei and Anandkumar, Anima",
    journal = "Nature Machine Intelligence",
    year = "2023",
}

@article{zhao2023gimlet,
    title = "{GIMLET}: A Unified Graph-Text Model for Instruction-Based Molecule Zero-Shot Learning",
    author = "Zhao, Haiteng and Liu, Shengchao and Ma, Chang and Xu, Hannan and Fu, Jie and Deng, Zhihong and Kong, Lingpeng and Liu, Qi",
    journal = "Advances in Neural Information Processing Systems",
    year = "2023",
}

@inproceedings{iupac1979,
  title={Nomenclature of organic chemistry: sections A, B, C, D, E, F and H - 1979 ed.},
  author={Jean Rigaudy and S. P. Klesney},
  year={1979},
  url={https://api.semanticscholar.org/CorpusID:109533074}
}

@article{zeng2022deep,
    title = "A Deep-learning System Bridging Molecule Structure and Biomedical Text with Comprehension Comparable to Human Professionals",
    author = "Zeng, Zheni and Yao, Yuan and Liu, Zhiyuan and Sun, Maosong",
    journal = "Nature Communications",
    year = "2022",
}

@article{chembl_web,
    author = {Davies, Mark and Nowotka, Michał and Papadatos, George and Dedman, Nathan and Gaulton, Anna and Atkinson, Francis and Bellis, Louisa and Overington, John P.},
    title = {ChEMBL web services: streamlining access to drug discovery data and utilities},
    journal = {Nucleic Acids Research},
    volume = {43},
    number = {W1},
    pages = {W612-W620},
    year = {2015},
    month = {07},
    issn = {0305-1048},
    doi = {10.1093/nar/gkv352},
    url = {https://doi.org/10.1093/nar/gkv352},
    eprint = {https://academic.oup.com/nar/article-pdf/43/W1/W612/17435802/gkv352.pdf},
}

@misc{fcd,
      title={Fr\'echet ChemNet Distance: A metric for generative models for molecules in drug discovery}, 
      author={Kristina Preuer and Philipp Renz and Thomas Unterthiner and Sepp Hochreiter and Günter Klambauer},
      year={2018},
      eprint={1803.09518},
      archivePrefix={arXiv},
      primaryClass={cs.LG},
      url={https://arxiv.org/abs/1803.09518}, 
}

@misc{moljson,
      title={Molecular Representations for Large Language Models}, 
      author={Nicholas T. Runcie and Fergus Imrie and Charlotte M. Deane},
      year={2026},
      eprint={2605.01822},
      archivePrefix={arXiv},
      primaryClass={cs.LG},
      url={https://arxiv.org/abs/2605.01822}, 
}

@article{hanwell2017open,
  title={Open chemistry: RESTful web APIs, JSON, NWChem and the modern web application},
  author={Marcus D. Hanwell and Wibe A. de Jong and Christopher J. Harris},
  journal={Journal of Cheminformatics},
  year={2017},
  volume={9},
  url={https://api.semanticscholar.org/CorpusID:3652559}
}

@software{commonchem,
  author    = {Matthew Swain},
  title     = {{CommonChem}: A Data Format for Chemical Information},
  url       = {https://github.com/CommonChem/CommonChem},
  urldate   = {2026-05-15},
  year = {2018}
}

@misc{mistral2025small24b,
  author       = {{Mistral AI}},
  title        = {{mistralai/Mistral-Small-24B-Instruct-2501}},
  year         = {2025},
  howpublished = {\url{https://huggingface.co/mistralai/Mistral-Small-24B-Instruct-2501}},
  note         = {Accessed: 2025-04-22}
}

@misc{ether0,
      title={Training a Scientific Reasoning Model for Chemistry}, 
      author={Siddharth M. Narayanan and James D. Braza and Ryan-Rhys Griffiths and Albert Bou and Geemi Wellawatte and Mayk Caldas Ramos and Ludovico Mitchener and Samuel G. Rodriques and Andrew D. White},
      year={2025},
      eprint={2506.17238},
      archivePrefix={arXiv},
      primaryClass={cs.LG},
      url={https://arxiv.org/abs/2506.17238}, 
}

@misc{anthropic2025haiku45,
  author       = {Anthropic},
  title        = {Introducing {Claude} {Haiku} 4.5},
  year         = {2025},
  month        = {October},
  howpublished = {\url{https://www.anthropic.com/news/claude-haiku-4-5}},
  note         = {Accessed: 2026-05-13}
}

@techreport{google2025gemini3flash,
  author      = {{Google DeepMind}},
  title       = {Gemini 3 Flash Model Card},
  institution = {Google DeepMind},
  year        = {2025},
  month       = dec,
  url         = {https://storage.googleapis.com/deepmind-media/Model-Cards/Gemini-3-Flash-Model-Card.pdf},
  note        = {Model released December 17, 2025}
}

@article{Bemis1996ThePO,
  title={The properties of known drugs. 1. Molecular frameworks.},
  author={Guy W. Bemis and Mark A. Murcko},
  journal={Journal of medicinal chemistry},
  year={1996},
  volume={39 15},
  pages={
          2887-93
        },
  url={https://api.semanticscholar.org/CorpusID:19424664}
}

@article{cml2000,
author = {Murray-Rust, Peter and Rzepa, Henry},
year = {2000},
month = {02},
pages = {},
title = {Chemical Markup, XML, and the Worldwide Web. 1. Basic Principles},
volume = {39},
journal = {Journal of Chemical Information and Computer Sciences},
doi = {10.1021/ci990052b}
}

@article{cml2003,
author = {Murray-Rust, Peter and Rzepa, Henry},
year = {2003},
month = {05},
pages = {757-72},
title = {Chemical markup, XML, and the world wide web. 4. CML schema},
volume = {43},
journal = {Journal of chemical information and computer sciences},
doi = {10.1021/ci0256541}
}

@article{tanimoto1957ibm,
  title={IBM internal report},
  author={Tanimoto, Taffee T},
  journal={Nov},
  volume={17},
  pages={1957},
  year={1957}
}

@inproceedings{liao-etal-2025-exploring,
    title = "Exploring Forgetting in Large Language Model Pre-Training",
    author = "Liao, Chonghua  and
      Xie, Ruobing  and
      Sun, Xingwu  and
      Sun, Haowen  and
      Kang, Zhanhui",
    editor = "Che, Wanxiang  and
      Nabende, Joyce  and
      Shutova, Ekaterina  and
      Pilehvar, Mohammad Taher",
    booktitle = "Proceedings of the 63rd Annual Meeting of the Association for Computational Linguistics (Volume 1: Long Papers)",
    month = jul,
    year = "2025",
    address = "Vienna, Austria",
    publisher = "Association for Computational Linguistics",
    url = "https://aclanthology.org/2025.acl-long.105/",
    doi = "10.18653/v1/2025.acl-long.105",
    pages = "2112--2127",
    ISBN = "979-8-89176-251-0"
}

@article{belinkov,
    author = {Belinkov, Yonatan},
    title = {Probing Classifiers: Promises, Shortcomings, and Advances},
    journal = {Computational Linguistics},
    volume = {48},
    number = {1},
    pages = {207-219},
    year = {2022},
    month = {04},
    issn = {0891-2017},
    doi = {10.1162/coli_a_00422},
    url = {https://doi.org/10.1162/coli_a_00422},
    eprint = {https://direct.mit.edu/coli/article-pdf/48/1/207/2006605/coli_a_00422.pdf},
}

@inproceedings{gao-etal-2024-insights,
    title = "Insights into {LLM} Long-Context Failures: When Transformers Know but Don{'}t Tell",
    author = "Gao, Muhan  and
      Lu, TaiMing  and
      Yu, Kuai  and
      Byerly, Adam  and
      Khashabi, Daniel",
    editor = "Al-Onaizan, Yaser  and
      Bansal, Mohit  and
      Chen, Yun-Nung",
    booktitle = "Findings of the Association for Computational Linguistics: EMNLP 2024",
    month = nov,
    year = "2024",
    address = "Miami, Florida, USA",
    publisher = "Association for Computational Linguistics",
    url = "https://aclanthology.org/2024.findings-emnlp.447/",
    doi = "10.18653/v1/2024.findings-emnlp.447",
    pages = "7611--7625"
}

@article{Kirkpatrick_2017,
   title={Overcoming catastrophic forgetting in neural networks},
   volume={114},
   ISSN={1091-6490},
   url={http://dx.doi.org/10.1073/pnas.1611835114},
   DOI={10.1073/pnas.1611835114},
   number={13},
   journal={Proceedings of the National Academy of Sciences},
   publisher={Proceedings of the National Academy of Sciences},
   author={Kirkpatrick, James and Pascanu, Razvan and Rabinowitz, Neil and Veness, Joel and Desjardins, Guillaume and Rusu, Andrei A. and Milan, Kieran and Quan, John and Ramalho, Tiago and Grabska-Barwinska, Agnieszka and Hassabis, Demis and Clopath, Claudia and Kumaran, Dharshan and Hadsell, Raia},
   year={2017},
   month=Mar, pages={3521–3526} }

@article{maccs,
  title={Reoptimization of MDL Keys for Use in Drug Discovery},
  author={Joseph L. Durant and Burton A. Leland and Douglas R. Henry and James G. Nourse},
  journal={Journal of chemical information and computer sciences},
  year={2002},
  volume={42 6},
  pages={
          1273-80
        },
  url={https://api.semanticscholar.org/CorpusID:22752474}
}

@article{morgan,
  title={Extended-Connectivity Fingerprints},
  author={David Rogers and Mathew Hahn},
  journal={Journal of chemical information and modeling},
  year={2010},
  volume={50 5},
  pages={
          742-54
        },
  url={https://api.semanticscholar.org/CorpusID:5132461}
}

@inproceedings{koehn-2004-statistical,
    title = "Statistical Significance Tests for Machine Translation Evaluation",
    author = "Koehn, Philipp",
    editor = "Lin, Dekang  and
      Wu, Dekai",
    booktitle = "Proceedings of the 2004 Conference on Empirical Methods in Natural Language Processing",
    month = jul,
    year = "2004",
    address = "Barcelona, Spain",
    publisher = "Association for Computational Linguistics",
    url = "https://aclanthology.org/W04-3250/",
    pages = "388--395"
}

@misc{yu2024llasmoladvancinglargelanguage,
      title={LlaSMol: Advancing Large Language Models for Chemistry with a Large-Scale, Comprehensive, High-Quality Instruction Tuning Dataset}, 
      author={Botao Yu and Frazier N. Baker and Ziqi Chen and Xia Ning and Huan Sun},
      year={2024},
      eprint={2402.09391},
      archivePrefix={arXiv},
      primaryClass={cs.AI},
      url={https://arxiv.org/abs/2402.09391}, 
}

@article{rdkit,
  author = {Landrum, Greg},
  description = {Release 2016_09_4 (Q3 2016) Release · rdkit/rdkit},
  title = {RDKit: Open-Source Cheminformatics Software},
  url = {https://github.com/rdkit/rdkit/releases/tag/Release_2016_09_4},
  year = 2016
}

@inproceedings{kwon2023efficient,
  title={Efficient Memory Management for Large Language Model Serving with PagedAttention},
  author={Woosuk Kwon and Zhuohan Li and Siyuan Zhuang and Ying Sheng and Lianmin Zheng and Cody Hao Yu and Joseph E. Gonzalez and Hao Zhang and Ion Stoica},
  booktitle={Proceedings of the ACM SIGOPS 29th Symposium on Operating Systems Principles},
  year={2023}
}

@article{yang2025qwen3,
  title  = {Qwen3 Technical Report},
  author = {Yang, An and Li, Anfeng and Yang, Baosong and Zhang, Beichen and Hui, Binyuan and Zheng, Bo and Yu, Bowen and Gao, Chang and Huang, Chengen and Lv, Chenxu and others},
  journal= {arXiv preprint arXiv:2505.09388},
  year   = {2025}
}

@article{abdin2024phi4,
  title  = {{Phi-4} Technical Report},
  author = {Abdin, Marah and Aneja, Jyoti and Behl, Harkirat and Bubeck, S{\'e}bastien and Eldan, Ronen and Gunasekar, Suriya and Harrison, Michael and Hewett, Russell J. and Javaheripi, Mojan and Kauffmann, Piero and others},
  journal= {arXiv preprint arXiv:2412.08905},
  year   = {2024}
}

@article{abdin2025phi4reasoning,
  title  = {{Phi-4-reasoning} Technical Report},
  author = {Abdin, Marah and Agarwal, Sahaj and Awadallah, Ahmed and Balachandran, Vidhisha and Behl, Harkirat and Chen, Lingjiao and de Rosa, Gustavo and Gunasekar, Suriya and Javaheripi, Mojan and Joshi, Neel and Kauffmann, Piero and Lara, Yash and Mendes, Caio C{\'e}sar Teodoro and Mitra, Arindam and Nushi, Besmira and Papailiopoulos, Dimitris and Saarikivi, Olli and Shah, Shital and Shrivastava, Vaishnavi and Vineet, Vibhav and Wu, Yue and Yousefi, Safoora and Zheng, Guoqing},
  journal= {arXiv preprint arXiv:2504.21318},
  year   = {2025}
}

@misc{olmo2025olmo3,
      title={Olmo 3}, 
      author={Team Olmo and : and Allyson Ettinger and Amanda Bertsch and Bailey Kuehl and David Graham and David Heineman and Dirk Groeneveld and Faeze Brahman and Finbarr Timbers and Hamish Ivison and Jacob Morrison and Jake Poznanski and Kyle Lo and Luca Soldaini and Matt Jordan and Mayee Chen and Michael Noukhovitch and Nathan Lambert and Pete Walsh and Pradeep Dasigi and Robert Berry and Saumya Malik and Saurabh Shah and Scott Geng and Shane Arora and Shashank Gupta and Taira Anderson and Teng Xiao and Tyler Murray and Tyler Romero and Victoria Graf and Akari Asai and Akshita Bhagia and Alexander Wettig and Alisa Liu and Aman Rangapur and Chloe Anastasiades and Costa Huang and Dustin Schwenk and Harsh Trivedi and Ian Magnusson and Jaron Lochner and Jiacheng Liu and Lester James V. Miranda and Maarten Sap and Malia Morgan and Michael Schmitz and Michal Guerquin and Michael Wilson and Regan Huff and Ronan Le Bras and Rui Xin and Rulin Shao and Sam Skjonsberg and Shannon Zejiang Shen and Shuyue Stella Li and Tucker Wilde and Valentina Pyatkin and Will Merrill and Yapei Chang and Yuling Gu and Zhiyuan Zeng and Ashish Sabharwal and Luke Zettlemoyer and Pang Wei Koh and Ali Farhadi and Noah A. Smith and Hannaneh Hajishirzi},
      year={2026},
      eprint={2512.13961},
      archivePrefix={arXiv},
      primaryClass={cs.CL},
      url={https://arxiv.org/abs/2512.13961}, 
}

@article{zhao2024chemdfm,
  title  = {Developing {ChemDFM} as a Large Language Foundation Model for Chemistry},
  author = {Zhao, Zihan and Ma, Da and Chen, Lu and Sun, Liangtai and Li, Zihao and Xia, Yi and Chen, Bo and Xu, Hongshen and Zhu, Zichen and Zhu, Su and Fan, Shuai and Shen, Guodong and Yu, Kai and Chen, Xin},
  journal= {Cell Reports Physical Science},
  year   = {2024},
  note   = {arXiv:2401.14818}
}

@article{zhao2025chemdfmr,
  title  = {{ChemDFM-R}: A Chemical Reasoning {LLM} Enhanced with Atomized Chemical Knowledge},
  author = {Zhao, Zihan and others},
  journal= {arXiv preprint arXiv:2507.21990},
  year   = {2025}
}

@misc{openai2026gpt54,
  title  = {Introducing {GPT-5.4}},
  author = {{OpenAI}},
  year   = {2026},
  month  = mar,
  note   = {Released March 5, 2026},
  url    = {https://openai.com/index/introducing-gpt-5-4/}
}

@article{pubchem2009,
  author = {Wang, Y. and Xiao, J. and Suzek, T. and Zhang, J. and Wang, J. and Bryant, S.},
  title = {PubChem: a public information system for analyzing bioactivities of small molecules},
  journal = {Nucleic Acids Research},
  year = {2009},
  volume = {37},
  number = {Web-Server-Issue},
  pages = {623-633}
}

@article{pneumocandin,
    author = {Denning, D W},
    title = {Echinocandins and pneumocandins--a new antifungal class with a novel mode of action.},
    journal = {Journal of Antimicrobial Chemotherapy},
    volume = {40},
    number = {5},
    pages = {611-614},
    year = {1997},
    month = {11},
    issn = {0305-7453},
    doi = {10.1093/jac/40.5.611},
    url = {https://doi.org/10.1093/jac/40.5.611},
    eprint = {https://academic.oup.com/jac/article-pdf/40/5/611/9837819/400611.pdf},
}

@article{qwen2025qwen25technicalreport,
  title={Qwen2.5 Technical Report},
  author={An Yang and Baosong Yang and Beichen Zhang and Binyuan Hui and Bo Zheng and Bowen Yu and Chengyuan Li and Dayiheng Liu and Fei Huang and Guanting Dong and Haoran Wei and Huan Lin and Jian Yang and Jianhong Tu and Jianwei Zhang and Jianxin Yang and Jiaxin Yang and Jingren Zhou and Junyang Lin and Kai Dang and Keming Lu and Keqin Bao and Kexin Yang and Le Yu and Mei Li and Mingfeng Xue and Pei Zhang and Qin Zhu and Rui Men and Runji Lin and Tianhao Li and Tingyu Xia and Xingzhang Ren and Xuancheng Ren and Yang Fan and Yang Su and Yi-Chao Zhang and Yunyang Wan and Yuqi Liu and Zeyu Cui and Zhenru Zhang and Zihan Qiu and Shanghaoran Quan and Zekun Wang},
  journal={ArXiv},
  year={2024},
  volume={abs/2412.15115},
  url={https://api.semanticscholar.org/CorpusID:274859421}
}

@article{zinc,
  title={ZINC: A Free Tool to Discover Chemistry for Biology},
  author={John J. Irwin and T. Sterling and Michael M. Mysinger and Erin S. Bolstad and Ryan G. Coleman},
  journal={Journal of Chemical Information and Modeling},
  year={2012},
  volume={52},
  pages={1757 - 1768},
  url={https://api.semanticscholar.org/CorpusID:9759396}
}

@book{wln,
  author    = {Wiswesser, William J.},
  title     = {A Line-Formula Chemical Notation},
  year      = {1954},
  publisher = {Crowell},
  address   = {New York}
}
\clearpage

\appendix

\section{Structured molecular representations}
\label{appendix_structured}

Structured representations of molecules are understudied in machine learning. 	JavaScript Object Notation (JSON) and Extensible Markup Language (XML) have been widely used for storing hierarchically structured data, providing schema-validated, human-readable, and programming language-agnostic storage formats. They have become widely used standards for data interchange on the internet and between scientific tools. In chemistry, JSON and XML formats have been developed for various purposes. Chemical Markup Language or CML \citep{cml2000} was the first domain-specific application of XML, encoding a molecule as nested elements, namely, \texttt{atomArray}, \texttt{bondArray}, and per-atom or per-bond attributes. This XML-based format makes every chemical component explicit and machine-parseable, so that chemical data could be represented or stored properly on web platforms. 

The cheminformatics community has shifted towards JSON for the same data interchange role, motivated by lighter parsers and native compatibility with web services and document stores. Chemical JSON \citep{hanwell2017open}, built based on CML, is the native representation of Avogadro 2. It stores a molecule as property arrays that map directly onto in-memory data structures and can be stored efficiently in binary JSON for database storage. Toolkit-oriented schemas such as CommonChem's JSON  \citep{commonchem} optimise for representation conversions between different cheminformatics libraries, while the ChemDoodle JSON representation \citep{chemdoodle} is meant for rendering molecular graphics on web browsers. Public databases like PubChem \citep{pubchem2009} and ChEMBL \citep{chembl_web} also expose molecules as JSON, though their schemas are optimized for retrieving data such as property tables and assay records. Recently, \citet{moljson} introduced another JSON representation for LLMs, MolJSON, and we investigate this in our benchmarking study.

\section{Model families}
\label{modelfamilies}

The open-weight models are served locally via vLLM \citep{kwon2023efficient} on H200 GPUs, while closed models are accessed via their respective API.  


\paragraph{Qwen3 (by Alibaba).}
The Qwen3 series \citep{yang2025qwen3} supports a switchable reasoning mode: \texttt{enable\_reasoning=True} activates chain-of-thought reasoning within \texttt{<think>} tags; \texttt{enable\_reasoning=False} gives direct responses. We evaluate Qwen3-4B (dense) and Qwen3-30B-A3B (Mixture-of-Experts model, 3B active parameters) under \emph{both} reasoning conditions, providing a clean within-model reasoning ablation. 

\paragraph{Phi-4 (by Microsoft).}
The Phi-4 family \citep{abdin2024phi4, abdin2025phi4reasoning} has three 14B models at different levels of reasoning: Phi-4, Phi-4-Reasoning (RL post-trained for reasoning), and Phi-4-Reasoning-Plus (stronger RL). 

\paragraph{OLMo-3 (by AI2).}
The OLMo-3 series \citep{olmo2025olmo3} provides fully open-weight 32B models. We include instruction-tuned OLMo-3.1-32B-Instruct and the reasoning variant OLMo-3.1-32B-Think.

\paragraph{Domain-specialized models.}
ChemDFM or Dialogue Foundation Model for Chemistry \citep{zhao2024chemdfm} is a chemistry-specialist LLM family which consists of ChemDFM-v2.0-14B and ChemDFM-R-14B \citep{zhao2025chemdfmr}. These models were obtained by finetuning Qwen2.5-14B \citep{qwen2025qwen25technicalreport} on chemical literature and molecular data. In addition, we also benchmark reasoning-capable Ether0 \citep{ether0} which is based on Mistral-Small-24B-Instruct-2501 \citep{mistral2025small24b}. We also benchmark the base LLMs of these domain-specialized models, namely Qwen2.5-14B and Mistral-Small-24B. Benchmarking such chemistry-specialized LLMs allows us to isolate the effect of domain-specialized LLMs from general LLMs.


\paragraph{Closed frontier models.}
GPT-5.4-Mini \citep{openai2026gpt54} and Claude-Haiku-4.5 \citep{anthropic2025haiku45} have been selected as closed-source frontier models for benchmarking. They were selected as they were two of the top mid-tier models by big players in the industry, namely, OpenAI and Anthropic. Their input token costs between \$0.75 to \$1.00 per 1M tokens based on which we could make a fair comparison and these versions allowed us to keep our API costs within the limited budget we were operating with.

\section{Source datasets}
\label{source_datasets}

ChEBI-20 pairs about 33,000 molecules with expert-written natural language descriptions and provides standard train/validation/test splits \citep{edwards-etal-2022-translation}. We use the test split as the source of molecules for our tasks. From the ChEBI-20 test split ($\sim$3.3K molecules), we select a stratified subset of 200 molecules for the comprehension benchmarks. We apply a complexity cut-off before sampling: molecules must have a molecular weight $\geq 300$\,Da and at least 2 rings. This cut-off excludes trivially simple molecules (e.g.\ small acyclic fragments, single amino acids) where all representations may succeed equally, ensuring that the benchmark discriminates between representations on structurally meaningful cases. 

For comprehension tasks (atom counting, functional group identification, property estimation), we use 200 molecules. For retrieval we use 78 molecules (each appearing as the correct answer once). Each retrieval question needs three distractor types (scaffold match, similar molecular weight, random). Not every molecule in the comprehension subset has another molecule in the test set that shares its Murcko scaffold \citep{Bemis1996ThePO}. Therefore, we ended up with a small dataset of 78 questions for this task, but it was sufficient to evaluate all models and draw clear insights. For isomer discrimination we construct 500 pairs. For caption-to-molecule generation, we use 200 molecules from the full test set.

ZINC250K is a curated, drug-like subset of the ZINC database \citep{zinc} containing 250,000 commercially available molecules. Unlike ChEBI-20, ZINC250K lacks natural language descriptions but provides a much larger pool of structurally diverse molecules. We use it to construct paired discrimination datasets (tautomer recognition and protonation state recognition) that require \emph{generating} chemically related variants of existing molecules as multiple choices for the questions. The large, diverse pool of drug-like molecules in ZINC250K helps with the generation of these variants.

\section{Tasks}
\label{appendix_data_tasks}

\paragraph{Atom Counting.}
Given a molecular string, the LLM is to count the number of atoms of a specified element (C, N, O, S, F, or Cl). The ground truth is computed from RDKit atom iteration. We measure performance by exact match accuracy.

\paragraph{Functional Group Identification.}
For each molecule, we prompt the LLM to identify the presence of different functional groups: aldehyde, ester, halide, primary amine, and sulfonamide. This is a binary classification task, where the LLM indicates the presence of each functional group with a \emph{yes} or a \emph{no}. The ground truth is detected via RDKit SMARTS matching, and the performance is measured by macro-averaged F1 across all 5 functional groups.

\paragraph{Molecular Property Estimation.}
The LLM is prompted to predict four different RDKit-computed properties when given each of the 9 representations separately. The properties are Wildman-Crippen LogP value, Topological Polar Surface Area and H-bond donor and acceptor counts, evaluated by exact match accuracy.

\paragraph{Molecule Retrieval.}
Given a ChEBI natural language description, identify the correct molecule from four candidates. This is a multiple-choice question task. The three distractors are selected based on the following factors: one shares the Murcko scaffold (same core, different substituents) \citep{Bemis1996ThePO}, one has a similar molecular weight ($\pm$10\%), and one is random. The metric used is top-1 accuracy.

\paragraph{Isomer Discrimination.}
The LLM, when given two molecular strings, has to determine whether they encode the same molecule or not, which is a binary classification task. We construct 500 pairs split across three types: positive pairs (same molecule but different atom orderings), stereoisomer pairs (molecules differing only in R/S or E/Z configuration), and substitution pairs (one atom changed). The stereoisomer pairs specifically probe whether representations that encode stereocenters help models detect chirality differences.

\paragraph{Tautomer Recognition.}
Here the task is to determine whether two given molecular strings are tautomeric forms of the same molecule. This is a binary classification task. Two molecules are tautomers if and only if they map to the same RDKit canonical tautomer. It contains 250 molecule pairs (125 positive: confirmed tautomeric forms; 125 negative: chemically confusable non-tautomers). Positive pairs span four tautomer classes: keto/enol, amide/imidic acid, heterocyclic, and nitroso/oxime, identified via RDKit's \texttt{TautomerEnumerator}. Negative pairs are unrelated molecules that do not share a canonical tautomer. 

\paragraph{Protonation State Recognition.}
Given two molecular strings from ZINC250K, the LLM has to determine whether they represent the same molecule at different protonation states, which is a binary classification task. It contains 210 pairs total. Two molecules are considered protonation variants if and only if their RDKit-neutralized canonical SMILES are identical. Negatives include molecules with similar charge patterns but genuinely different structures. This task contains 210 molecule pairs (105 positive pairs, which are same molecule at different protonation states and 105 negative pairs). Positive pairs cover carboxylic acid/carboxylate, amine/ammonium, phenol/phenolate, and zwitterion variants generated via RDKit's \texttt{MolStandardize} module \citep{rdkit}. Ground truth uses RDKit's \texttt{Uncharger}-based neutralization: two molecules are protonation variants if and only if they share the same neutral canonical SMILES.

\paragraph{Caption-to-Molecule Generation.}
This task involves generating the molecular string in the target representation when given a natural language molecular description or caption from the ChEBI-20 dataset.  We use the following metrics for evaluation: validity rate (fraction of the generated molecular strings that are parseable to a valid RDKit molecule), exact match (canonical SMILES of generated molecule matches ground truth), Tanimoto similarity (2048-bit Morgan fingerprint with radius 2), and Fréchet ChemNet Distance or FCD \citep{fcd}.

\section{Prompt templates}
\label{app:prompts}

This section lists the exact prompt templates used for each benchmark task.
Placeholders are shown in \texttt{\{curly braces\}}.
All models receive the same task prompt; the only variation across conditions is whether the model's reasoning/reasoning mode is enabled (controlled via chat template parameters or system messages, not via the prompt itself).
For the generation benchmark, we use the structured output capability supported by vLLM (Section \ref{app:prompt-b6}).

\subsection{Atom Counting}
\label{app:prompt-b1}

\begin{promptbox}[Prompt - Atom Counting]
How many \{element\} atoms are in the following molecule?\newline
Molecule: \{molecule\_str\}\newline
\newline
Please reason step by step, and put your final answer within \textbackslash boxed\{\}.\newline
\newline
Example format: After reasoning, conclude with: The answer is \textbackslash boxed\{42\}
\end{promptbox}

\vars{\texttt{\{element\}} $\in$ \{C, N, O, S, F, Cl\};\quad
\texttt{\{molecule\_str\}}: molecule in the target representation.}

\subsection{Functional Group Identification}
\label{app:prompt-b2}

\begin{promptbox}[Prompt - Functional Group Identification]
Does the following molecule contain a \{functional\_group\}?\newline
Molecule: \{molecule\_str\}\newline
\newline
Please reason step by step, and put your final answer (Yes or No) within \textbackslash boxed\{\}.\newline
\newline
Example format: After reasoning, conclude with: The answer is \textbackslash boxed\{Yes\} or \textbackslash boxed\{No\}
\end{promptbox}

\vars{\texttt{\{functional\_group\}} $\in$ \{Primary amine, Ester, Aldehyde, Sulfonamide, Halide\};\quad
\texttt{\{molecule\_str\}}: molecule in the target representation.}

\subsection{Property estimation}
\label{app:prompt-b3}

\begin{promptbox}[Prompt - Property estimation]
Estimate the \{property\_display\} of the following molecule.\newline
Molecule: \{molecule\_str\}\newline
\newline
Please reason step by step, and put your final answer (a number) within \textbackslash boxed\{\}.\newline
\newline
Example format: After reasoning, conclude with: The answer is \textbackslash boxed\{3.5\}
\end{promptbox}

\vars{\texttt{\{property\_display\}} $\in$ \{LogP (partition coefficient), Topological Polar Surface Area (TPSA), number of hydrogen bond donors, number of hydrogen bond acceptors\};\quad
\texttt{\{molecule\_str\}}: molecule in the target representation.}

\subsection{Molecule Retrieval}
\label{app:prompt-b4}

\begin{promptbox}[Prompt - Molecule Retrieval]
Which of the following molecules matches this description?\newline
Description: \{description\}\newline
A: \{molecule\_A\}\newline
B: \{molecule\_B\}\newline
C: \{molecule\_C\}\newline
D: \{molecule\_D\}\newline
\newline
Please reason step by step, and put your final answer (only the choice letter) within \textbackslash boxed\{\}.\newline
\newline
Example format: After reasoning, conclude with: The answer is \textbackslash boxed\{A\}
\end{promptbox}

\vars{\texttt{\{description\}}: natural language molecule description from ChEBI-20;\quad
\texttt{\{molecule\_A\}}-\texttt{\{molecule\_D\}}: four candidate molecules in the target representation (one correct, three distractors).}

\subsection{Isomer Discrimination}
\label{app:prompt-b5}

\begin{promptbox}[Prompt - Isomer Discrimination]
Do the following two molecular representations refer to the same molecule?\newline
Molecule 1: \{molecule\_1\}\newline
Molecule 2: \{molecule\_2\}\newline
\newline
Please reason step by step, and put your final answer (Yes or No) within \textbackslash boxed\{\}.\newline
\newline
Example format: After reasoning, conclude with: The answer is \textbackslash boxed\{Yes\} or \textbackslash boxed\{No\}
\end{promptbox}

\vars{Pairs are drawn from three categories: identical molecules (positive), constitutional isomers (negative-substitution), and stereoisomers (negative-stereoisomer).}

\subsection{Caption-to-Molecule Generation}
\label{app:prompt-b6}

\begin{promptbox}[Prompt - Caption-to-Molecule Generation]
Given a description of a molecule, generate the corresponding \{rep\_name\} string.\newline
\newline
IMPORTANT: Don't restate the full description - get straight to generating the molecule. Keep any reasoning brief and concise. Output tokens are limited, so provide your final answer as quickly as possible.\newline
\newline
Description: \{description\}\newline
\newline
Provide your final answer in the format: Final answer: <molecule\_string>
\end{promptbox}

\begin{promptbox}[Prompt - MolJSON instruction]
Your final output must be ONLY a valid JSON object with "atoms" and "bonds" arrays.\newline
\newline
Format:\newline
\{"atoms": [\{"id": "C1", "element": "C"\}, ...],\newline
\ "bonds": [\{"source": "C1", "target": "C2", "order": 1.0\}, ...],\newline
\ "charges": null, "aromatic\_n\_h": null\}\newline
\newline
Keep the reasoning concise and short. Don't ramble, be decisive. As soon as possible, output the JSON object.
\end{promptbox}

We also provide LLMs the schema for MolJSON provided by \citet{moljson} using vLLM's \texttt{StructuredOutputsParams} for constraining the model's output to a valid format \citep{kwon2023efficient}. 

\begin{promptbox}[Prompt - JSON formatting for non-MolJSON]
Your final output must be a JSON object in this format:\newline
\{"molecule": "..."\}\newline
\newline
Where the value is the complete \{rep\_name\} string. Keep the reasoning concise and short. Don't ramble, be decisive. As soon as possible, output the JSON object.
\end{promptbox}

\subsection{Tautomer Recognition}
\label{app:prompt-b9}

\begin{promptbox}[Prompt - Tautomer Recognition]
Are the following two molecules tautomeric forms of the same molecule?\newline
Molecule 1: \{molecule\_1\}\newline
Molecule 2: \{molecule\_2\}\newline
\newline
Please reason step by step, and put your final answer (Yes or No) within \textbackslash boxed\{\}.\newline
\newline
Example format: After reasoning, conclude with: The answer is \textbackslash boxed\{Yes\} or \textbackslash boxed\{No\}
\end{promptbox}

\subsection{Protonation State Recognition}
\label{app:prompt-b10}

\begin{promptbox}[Prompt - Protonation State Recognition]
Are the following two molecules different protonation states of the same molecule?\newline
Molecule 1: \{molecule\_1\}\newline
Molecule 2: \{molecule\_2\}\newline
\newline
Please reason step by step, and put your final answer (Yes or No) within \textbackslash boxed\{\}.\newline
\newline
Example format: After reasoning, conclude with: The answer is \textbackslash boxed\{Yes\} or \textbackslash boxed\{No\}
\end{promptbox}


\section{Ether0's performance across representations}
\label{ether0_app}
Ether0-24B is post-trained from Mistral-Small-24B-Instruct-2501 for chemistry, yet Mistral-Small-24B outperforms Ether0 on essentially every task in our benchmark. On property estimation, Ether0 scores 0.102 in accuracy with canonical SMILES versus Mistral's 0.444. On molecule retrieval, 0.269 versus 0.513. On atom counting, 0.195 versus 0.360. On tautomer and protonation recognition Ether0 is at or below the 0.50 random baseline on every representation, while Mistral exceeds 0.95 accuracy on canonical SMILES, isomeric SMILES, MolJSON, and InChI for tautomer; and 0.99 for protonation. Most strikingly, Ether0 fails caption-to-molecule generation across the board: validity rates are 0.00 to 0.20 (Table \ref{tab:caption_to_molecule_validity_rate}) and exact match is uniformly 0.00 (Table \ref{tab:caption_to_molecule_exact_match}). This underperformance by Ether0 is surprising as its post-training involves tasks relevant to ours, such as IUPAC name to SMILES translation, molecular captioning with molecules from \citet{yu2024llasmoladvancinglargelanguage}, and functional group suggestion \citep{ether0}. The potential reasons could be that Ether0's post-training might have eroded its general molecular reasoning, which is exhibited somewhat by its base model, and it is struggling  to generalize beyond its
training distribution as noted by \citet{ether0}.

\section{Reasoning is not consistently beneficial}
\label{reasoning_appendix}



The model families we have selected help to analyse the effect of reasoning mode from three different angles: (1) between the reasoning and non-reasoning modes of the Qwen3 models, (2) the Phi-4 models, which are all reasoning models but with progressively more reasoning supervision and, (3) the OLMo-3.1 models, which have instruction-tuned and reasoning variants.

The Qwen3 differences between their reasoning and non-reasoning variants are quite small. Qwen3-30B atom counting on canonical SMILES moves from 0.855 (non-reasoning) to 0.805 (reasoning), within confidence intervals. OLMo3.1-32B-Think vs. Instruct is largely positive on generation (MACCS tanimoto similarity with IUPAC improves from 0.312 to 0.511) but regresses on tautomer recognition across every SMILES variant (for example, with canonical SMILES it drops from 0.744  to 0.620), suggesting reasoning-style training can push the model away from competent recognition performance.

The Phi-4 family's reasoning progression also reveals some inconsistent patterns. From Phi-4 to Phi-4-Reasoning, atom counting with canonical SMILES jumps from 0.425 to 0.715 in accuracy. In TPSA property estimation, with MolJSON and InChI we see significant progress but the improvements are more stagnant with the SMILES variants. In HBD counting, there is improvement from Phi-4 to Phi-4 Reasoning across all representations but from Phi-4-Reasoning to Reasoning-Plus, there are significant drops which may point to the heavier post-training resulting in catastrophic forgetting \citep{Kirkpatrick_2017, liao-etal-2025-exploring}. However, across the three models from Phi-4 to Phi-4-Reasoning Plus, on caption-to-molecule generation with canonical SMILES, the validity rate drops from 0.712 to 0.324 to 0.356 (Table \ref{tab:caption_to_molecule_validity_rate}) while exact match rises incrementally from 0.016 to 0.048 to 0.056 (Table \ref{tab:caption_to_molecule_exact_match}).

Therefore, across model families, enabling reasoning mode does not uniformly improve performance. In several cases, reasoning variants perform similarly to or worse than their non-reasoning counterparts. However, in the case of molecule generation, reasoning models help cut down on hallucination (Figure \ref{qual_b6_hallucination_drop}) as we investigate in Section \ref{llmjudge}.
\clearpage
\section{Tokenization audit diagrams}
\begin{figure}[ht]
  \centering
  \includegraphics[width=0.9\columnwidth]{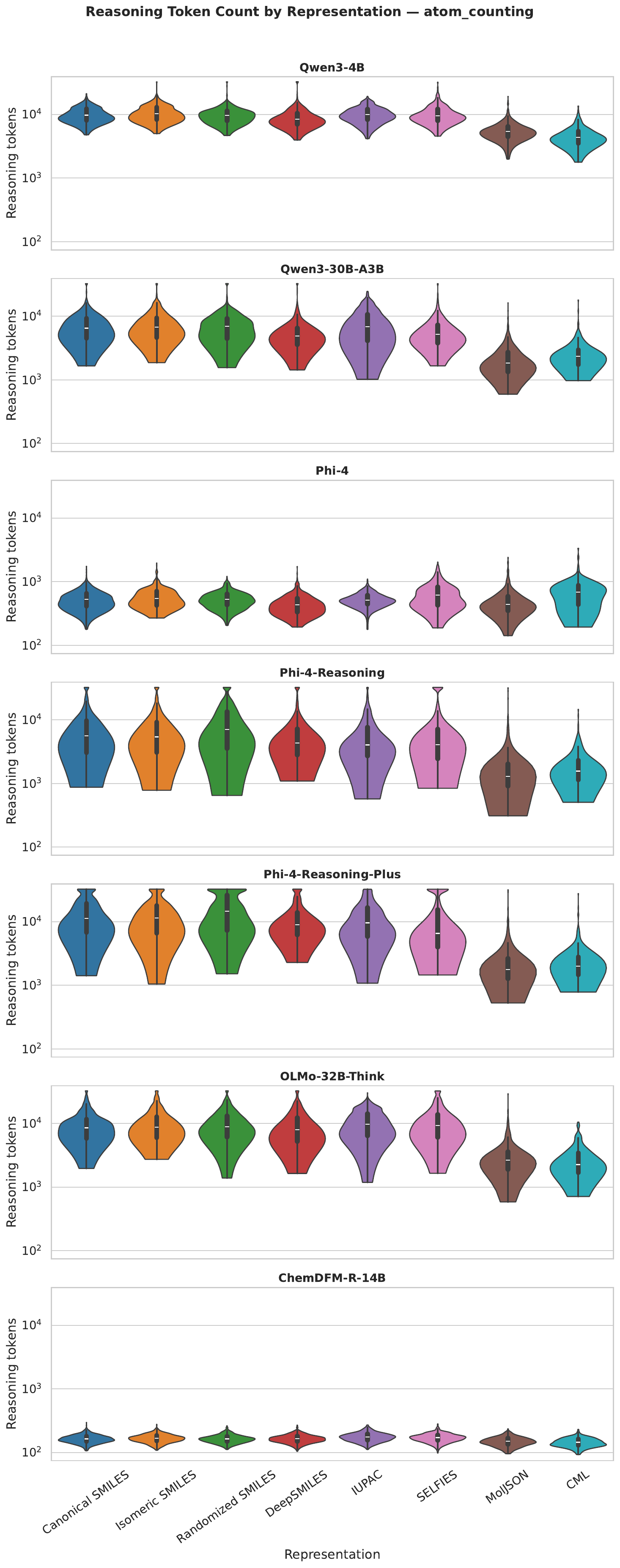}
  \caption{
    Reasoning token lengths for atom counting task
  }
  \label{reasoning_len__atom_counting}
\end{figure}

\section{Attention analysis}
\label{appendix_attention}



\begin{figure}[H]
  \centering
  \includegraphics[width=\columnwidth]{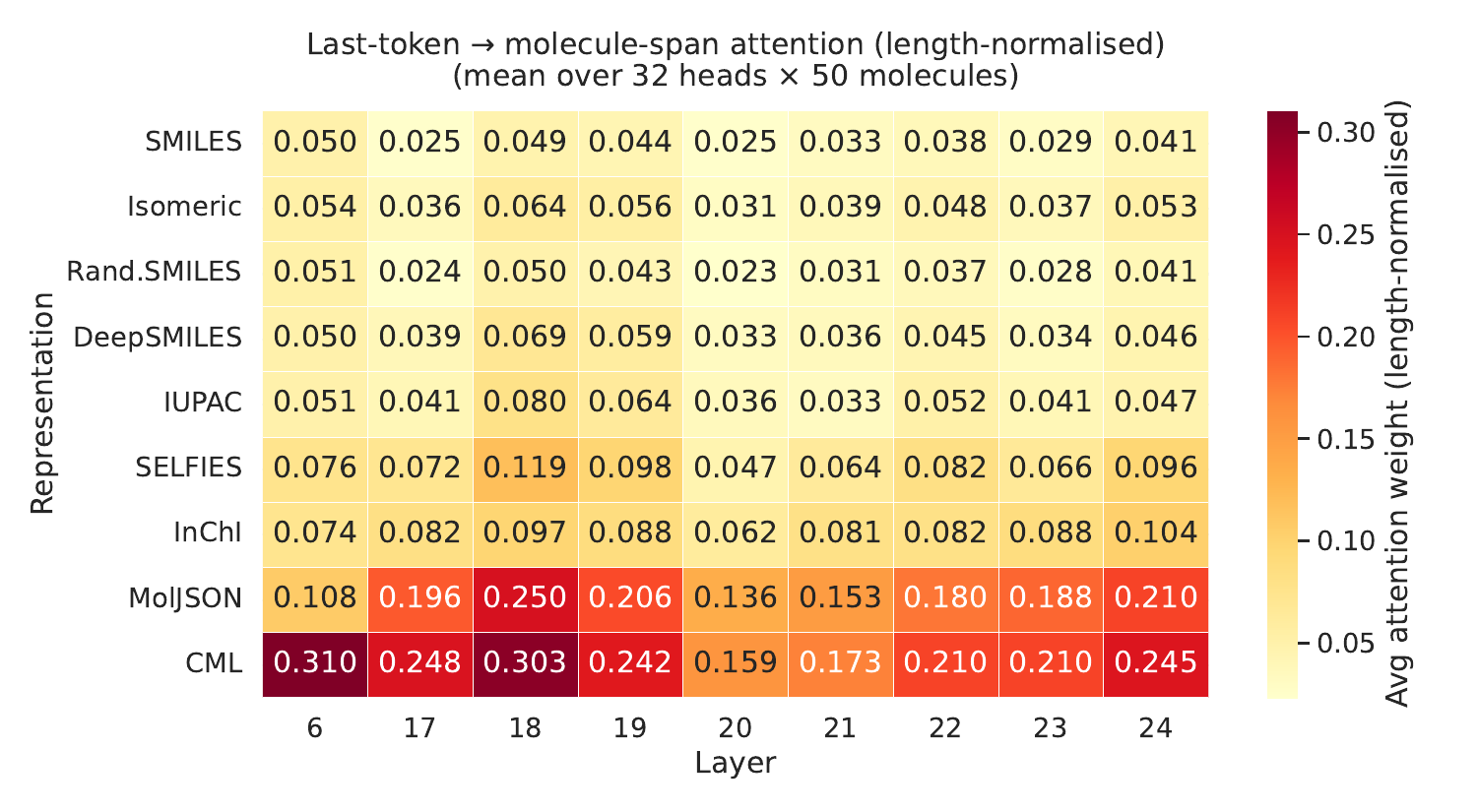}
  \caption{
    Attention between the last token and the molecule (length-normalised)
  }
  \label{last-tok-mol}
\end{figure}

\begin{figure}[H]
  \centering
  \includegraphics[width=\columnwidth]{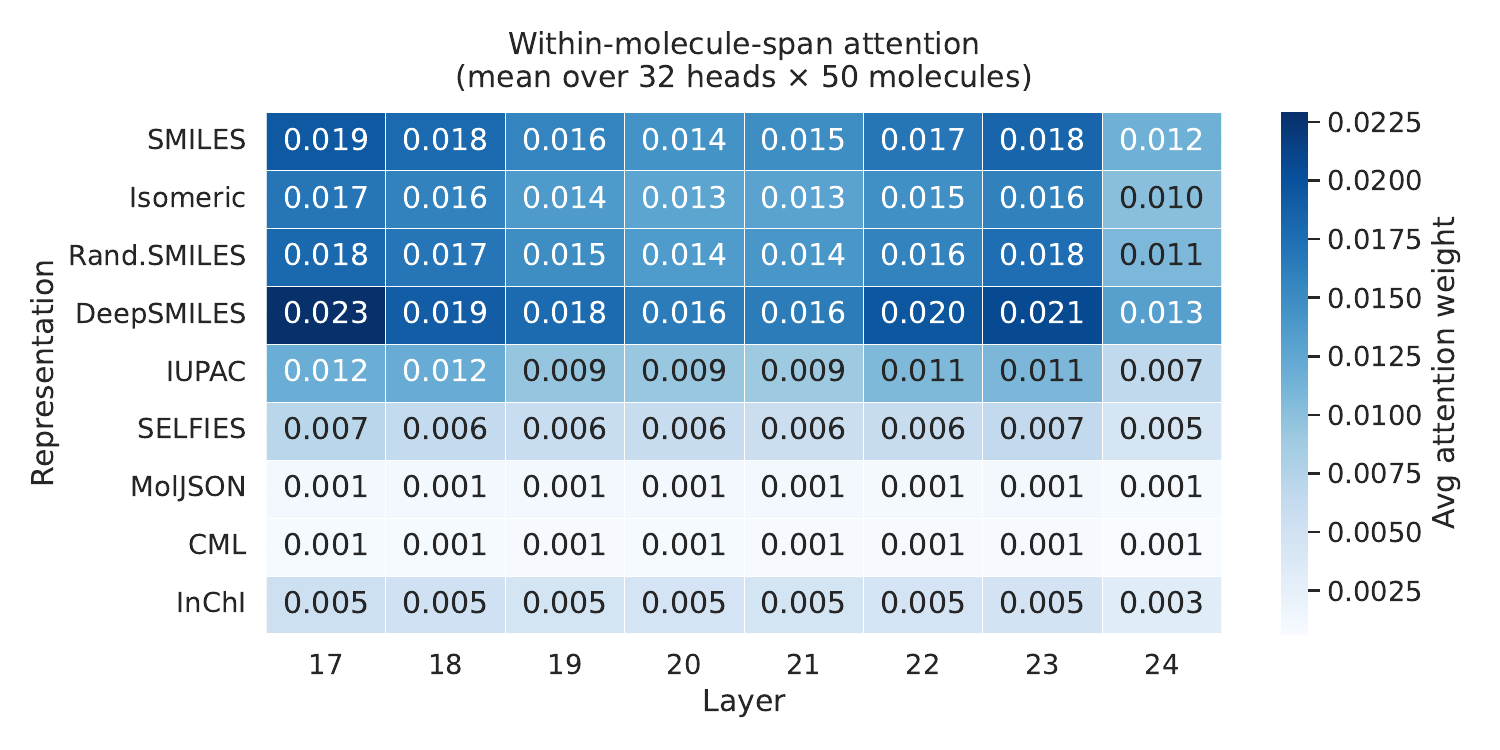}
  \caption{
    Attention within the molecule
  }
  \label{within-mol}
\end{figure}

\onecolumn

\section{Molecular text representations}
\label{fig:appendix_repr_examples}

This section lists the nine molecular representations used in this work, illustrated for L-Alanine (C$_3$H$_7$NO$_2$).
\definecolor{canonicalsmiles}{RGB}{31,119,180}
\definecolor{isomericsmiles}{RGB}{255,127,14}
\definecolor{randomizedsmiles}{RGB}{44,160,44}
\definecolor{deepsmiles}{RGB}{214,39,40}
\definecolor{iupac}{RGB}{148,103,189}
\definecolor{selfies}{RGB}{227,119,194}
\definecolor{moljson}{RGB}{140,86,75}
\definecolor{cml}{RGB}{23,190,207}
\definecolor{inchi}{RGB}{127,127,127}

\begin{figure}[H]
\centering
\includegraphics[width=0.4\columnwidth]{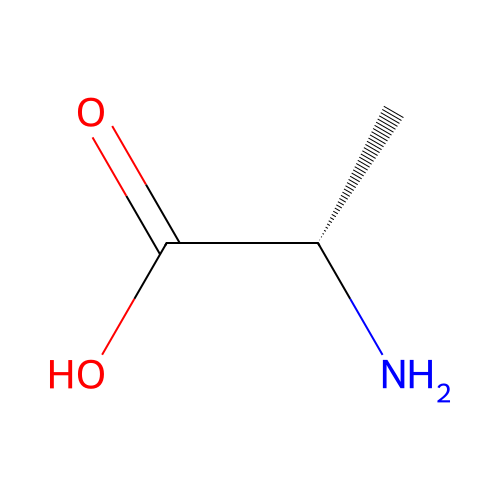}
\vspace{0.2em}

\begin{tcolorbox}[colback=canonicalsmiles!8!white,colframe=canonicalsmiles,coltitle=white,fonttitle=\bfseries\scriptsize,title={Canonical SMILES},left=1pt,right=1pt,top=0pt,bottom=0pt,boxsep=1pt,sharp corners,boxrule=0.5pt,toptitle=1pt,bottomtitle=1pt]
{\ttfamily\tiny CC(N)C(=O)O}
\end{tcolorbox}
\vspace{-0.5em}

\begin{tcolorbox}[colback=isomericsmiles!8!white,colframe=isomericsmiles,coltitle=white,fonttitle=\bfseries\scriptsize,title={Isomeric SMILES},left=1pt,right=1pt,top=0pt,bottom=0pt,boxsep=1pt,sharp corners,boxrule=0.5pt,toptitle=1pt,bottomtitle=1pt]
{\ttfamily\tiny C[C@@H](N)C(=O)O}
\end{tcolorbox}
\vspace{-0.5em}

\begin{tcolorbox}[colback=randomizedsmiles!8!white,colframe=randomizedsmiles,coltitle=white,fonttitle=\bfseries\scriptsize,title={Randomized SMILES},left=1pt,right=1pt,top=0pt,bottom=0pt,boxsep=1pt,sharp corners,boxrule=0.5pt,toptitle=1pt,bottomtitle=1pt]
{\ttfamily\tiny C(C(=O)O)(N)C}
\end{tcolorbox}
\vspace{-0.5em}

\begin{tcolorbox}[colback=deepsmiles!8!white,colframe=deepsmiles,coltitle=white,fonttitle=\bfseries\scriptsize,title={DeepSMILES},left=1pt,right=1pt,top=0pt,bottom=0pt,boxsep=1pt,sharp corners,boxrule=0.5pt,toptitle=1pt,bottomtitle=1pt]
{\ttfamily\tiny CCN)C=O)O}
\end{tcolorbox}
\vspace{-0.5em}

\begin{tcolorbox}[colback=iupac!8!white,colframe=iupac,coltitle=white,fonttitle=\bfseries\scriptsize,title={IUPAC},left=1pt,right=1pt,top=0pt,bottom=0pt,boxsep=1pt,sharp corners,boxrule=0.5pt,toptitle=1pt,bottomtitle=1pt]
{\ttfamily\tiny (2S)-2-aminopropanoic acid}
\end{tcolorbox}
\vspace{-0.5em}

\begin{tcolorbox}[colback=selfies!8!white,colframe=selfies,coltitle=white,fonttitle=\bfseries\scriptsize,title={SELFIES},left=1pt,right=1pt,top=0pt,bottom=0pt,boxsep=1pt,sharp corners,boxrule=0.5pt,toptitle=1pt,bottomtitle=1pt]
{\ttfamily\tiny [C][C][Branch1][C][N][C][=Branch1][C][=O][O]}
\end{tcolorbox}
\vspace{-0.5em}

\begin{tcolorbox}[colback=moljson!8!white,colframe=moljson,coltitle=white,fonttitle=\bfseries\scriptsize,title={MolJSON},left=1pt,right=1pt,top=0pt,bottom=0pt,boxsep=1pt,sharp corners,boxrule=0.5pt,toptitle=1pt,bottomtitle=1pt]
\begin{lstlisting}[basicstyle=\ttfamily\tiny,breaklines=true,breakautoindent=false,columns=fullflexible,aboveskip=0pt,belowskip=0pt]
{"atoms": [
  {"id":"C1","element":"C"}, {"id":"C2","element":"C"}, {"id":"N1","element":"N"}, {"id":"C3","element":"C"}, {"id":"O1","element":"O"}, {"id":"O2","element":"O"}
],
"bonds": [
  {"source":"C1","target":"C2","order":1.0}, {"source":"C2","target":"N1","order":1.0}, {"source":"C2","target":"C3","order":1.0}, {"source":"C3","target":"O1","order":2.0}, {"source":"C3","target":"O2","order":1.0}
]}
\end{lstlisting}
\end{tcolorbox}
\vspace{-0.5em}

\begin{tcolorbox}[colback=cml!8!white,colframe=cml,coltitle=white,fonttitle=\bfseries\scriptsize,title={CML},left=1pt,right=1pt,top=0pt,bottom=0pt,boxsep=1pt,sharp corners,boxrule=0.5pt,toptitle=1pt,bottomtitle=1pt]
\begin{lstlisting}[basicstyle=\ttfamily\tiny,breaklines=true,breakautoindent=false,columns=fullflexible,aboveskip=0pt,belowskip=0pt]
<molecule spinMultiplicity="2">
 <atomArray>
  <atom id="a1" elementType="C" hydrogenCount="3"/>
  <atom id="a2" elementType="C" hydrogenCount="1"/>
  <atom id="a3" elementType="N" hydrogenCount="2"/>
  <atom id="a4" elementType="C" hydrogenCount="0"/>
  <atom id="a5" elementType="O" hydrogenCount="0"/>
  <atom id="a6" elementType="O" hydrogenCount="1"/>
 </atomArray>
 <bondArray>
  <bond atomRefs2="a1 a2" order="1"/>
  <bond atomRefs2="a2 a3" order="1"/>
  <bond atomRefs2="a2 a4" order="1"/>
  <bond atomRefs2="a4 a5" order="2"/>
  <bond atomRefs2="a4 a6" order="1"/>
 </bondArray>
</molecule>
\end{lstlisting}
\end{tcolorbox}
\vspace{-0.5em}

\begin{tcolorbox}[colback=inchi!8!white,colframe=inchi,coltitle=white,fonttitle=\bfseries\scriptsize,title={InChI},left=1pt,right=1pt,top=0pt,bottom=0pt,boxsep=1pt,sharp corners,boxrule=0.5pt,toptitle=1pt,bottomtitle=1pt]
{\ttfamily\tiny InChI=1S/C3H7NO2/c1-2(4)3(5)6/h2H,4H2,1H3,(H,5,6)}
\end{tcolorbox}
\label{fig:repr_examples}
\end{figure}

\clearpage

\onecolumn
\section{Tables of results for all tasks in MolRepBench}
\label{appendix_tables_results}
This section contains the full results in Tables \ref{tab:atom_counting_accuracy} to \ref{tab:protonation_state_recognition_accuracy}.

\begin{table}[H]
\centering
\resizebox{\textwidth}{!}{
\begin{tabular}{llccccccccc}
\toprule
Model & Reasoning & Canonical SMILES & Isomeric SMILES & Randomized SMILES & DeepSMILES & IUPAC & SELFIES & MolJSON & CML & InChI \\
\midrule
Qwen3-4B & \checkmark & 0.740{\scriptsize$\pm$0.031} & 0.715{\scriptsize$\pm$0.032} & 0.690{\scriptsize$\pm$0.033} & 0.700{\scriptsize$\pm$0.032} & 0.710{\scriptsize$\pm$0.032} & 0.365{\scriptsize$\pm$0.034} & 0.800{\scriptsize$\pm$0.028} & \textbf{0.995{\scriptsize$\pm$0.005}} & 0.955{\scriptsize$\pm$0.015} \\
Qwen3-4B & \texttimes & 0.740{\scriptsize$\pm$0.031} & 0.795{\scriptsize$\pm$0.029} & 0.660{\scriptsize$\pm$0.034} & 0.695{\scriptsize$\pm$0.033} & 0.700{\scriptsize$\pm$0.032} & 0.375{\scriptsize$\pm$0.034} & 0.830{\scriptsize$\pm$0.027} & \textbf{0.995{\scriptsize$\pm$0.005}} & 0.945{\scriptsize$\pm$0.016} \\
Qwen3-30B & \checkmark & 0.805{\scriptsize$\pm$0.028} & \textbf{0.820{\scriptsize$\pm$0.027}} & 0.735{\scriptsize$\pm$0.031} & 0.885{\scriptsize$\pm$0.023} & 0.780{\scriptsize$\pm$0.029} & 0.410{\scriptsize$\pm$0.035} & 0.990{\scriptsize$\pm$0.007} & \textbf{0.995{\scriptsize$\pm$0.005}} & 0.995{\scriptsize$\pm$0.005} \\
Qwen3-30B & \texttimes & \textbf{0.855{\scriptsize$\pm$0.025}} & 0.810{\scriptsize$\pm$0.028} & 0.740{\scriptsize$\pm$0.031} & 0.850{\scriptsize$\pm$0.025} & 0.785{\scriptsize$\pm$0.029} & 0.425{\scriptsize$\pm$0.035} & 0.990{\scriptsize$\pm$0.007} & \textbf{0.995{\scriptsize$\pm$0.005}} & 0.995{\scriptsize$\pm$0.005} \\
Phi-4 & \checkmark & 0.425{\scriptsize$\pm$0.035} & 0.445{\scriptsize$\pm$0.035} & 0.395{\scriptsize$\pm$0.035} & 0.395{\scriptsize$\pm$0.035} & 0.515{\scriptsize$\pm$0.035} & 0.255{\scriptsize$\pm$0.031} & 0.925{\scriptsize$\pm$0.019} & 0.720{\scriptsize$\pm$0.032} & \textbf{1.000} \\
\bottomrule
\end{tabular}}
\caption{Atom Counting (accuracy)}
\label{tab:atom_counting_accuracy}
\end{table}

\begin{table}[H]
\centering
\resizebox{\textwidth}{!}{\begin{tabular}{llccccccccc}
\toprule
Model & Reasoning & Canonical SMILES & Isomeric SMILES & Randomized SMILES & DeepSMILES & IUPAC & SELFIES & MolJSON & CML & InChI \\
\midrule
Qwen3-4B & \checkmark & 0.583{\scriptsize$\pm$0.054} & 0.676{\scriptsize$\pm$0.066} & 0.530{\scriptsize$\pm$0.048} & 0.579{\scriptsize$\pm$0.022} & 0.719{\scriptsize$\pm$0.082} & 0.262{\scriptsize$\pm$0.020} & 0.680{\scriptsize$\pm$0.043} & 0.765{\scriptsize$\pm$0.078} & 0.348{\scriptsize$\pm$0.023} \\
Qwen3-4B & \texttimes & 0.591{\scriptsize$\pm$0.037} & 0.596{\scriptsize$\pm$0.033} & 0.545{\scriptsize$\pm$0.034} & 0.566{\scriptsize$\pm$0.030} & 0.729{\scriptsize$\pm$0.078} & 0.322{\scriptsize$\pm$0.053} & 0.647{\scriptsize$\pm$0.047} & 0.847{\scriptsize$\pm$0.075} & 0.338{\scriptsize$\pm$0.025} \\
Qwen3-30B & \checkmark & 0.612{\scriptsize$\pm$0.037} & 0.837{\scriptsize$\pm$0.074} & 0.590{\scriptsize$\pm$0.033} & 0.587{\scriptsize$\pm$0.033} & 0.720{\scriptsize$\pm$0.063} & 0.293{\scriptsize$\pm$0.020} & 0.741{\scriptsize$\pm$0.034} & 0.827{\scriptsize$\pm$0.058} & 0.401{\scriptsize$\pm$0.025} \\
Qwen3-30B & \texttimes & 0.806{\scriptsize$\pm$0.080} & 0.833{\scriptsize$\pm$0.075} & 0.684{\scriptsize$\pm$0.064} & 0.597{\scriptsize$\pm$0.028} & 0.743{\scriptsize$\pm$0.062} & 0.343{\scriptsize$\pm$0.053} & 0.730{\scriptsize$\pm$0.029} & \textbf{0.854{\scriptsize$\pm$0.066}} & 0.418{\scriptsize$\pm$0.027} \\
Phi-4 & \checkmark & 0.347{\scriptsize$\pm$0.032} & 0.448{\scriptsize$\pm$0.058} & 0.332{\scriptsize$\pm$0.029} & 0.313{\scriptsize$\pm$0.033} & 0.665{\scriptsize$\pm$0.072} & 0.276{\scriptsize$\pm$0.028} & 0.470{\scriptsize$\pm$0.034} & 0.479{\scriptsize$\pm$0.035} & 0.294{\scriptsize$\pm$0.022} \\
\bottomrule
\end{tabular}}
\caption{Functional Groups (Macro F1)}
\label{tab:functional_groups_macro_f1}
\end{table}

\begin{table}[H]
\centering
\resizebox{\textwidth}{!}{\begin{tabular}{llccccccccc}
\toprule
Model & Reasoning & Canonical SMILES & Isomeric SMILES & Randomized SMILES & DeepSMILES & IUPAC & SELFIES & MolJSON & CML & InChI \\
\midrule
Qwen3-4B & \checkmark & 0.616{\scriptsize$\pm$0.051} & 0.629{\scriptsize$\pm$0.052} & 0.547{\scriptsize$\pm$0.058} & 0.369{\scriptsize$\pm$0.071} & 0.709{\scriptsize$\pm$0.041} & 0.427{\scriptsize$\pm$0.063} & 0.589{\scriptsize$\pm$0.053} & 0.570{\scriptsize$\pm$0.057} & 0.740{\scriptsize$\pm$0.039} \\
Qwen3-4B & \texttimes & 0.677{\scriptsize$\pm$0.046} & 0.643{\scriptsize$\pm$0.050} & 0.559{\scriptsize$\pm$0.054} & 0.320{\scriptsize$\pm$0.073} & 0.727{\scriptsize$\pm$0.048} & 0.281{\scriptsize$\pm$0.071} & 0.561{\scriptsize$\pm$0.059} & 0.496{\scriptsize$\pm$0.066} & 0.742{\scriptsize$\pm$0.033} \\
Qwen3-30B & \checkmark & 0.710{\scriptsize$\pm$0.046} & 0.737{\scriptsize$\pm$0.040} & 0.713{\scriptsize$\pm$0.040} & 0.583{\scriptsize$\pm$0.058} & 0.720{\scriptsize$\pm$0.042} & 0.534{\scriptsize$\pm$0.056} & 0.682{\scriptsize$\pm$0.047} & 0.632{\scriptsize$\pm$0.053} & 0.743{\scriptsize$\pm$0.040} \\
Qwen3-30B & \texttimes & 0.698{\scriptsize$\pm$0.045} & 0.723{\scriptsize$\pm$0.042} & 0.669{\scriptsize$\pm$0.044} & 0.643{\scriptsize$\pm$0.049} & 0.697{\scriptsize$\pm$0.047} & 0.513{\scriptsize$\pm$0.058} & 0.654{\scriptsize$\pm$0.050} & 0.674{\scriptsize$\pm$0.044} & 0.732{\scriptsize$\pm$0.042} \\
Phi-4 & \checkmark & 0.661{\scriptsize$\pm$0.050} & 0.705{\scriptsize$\pm$0.043} & 0.660{\scriptsize$\pm$0.048} & 0.412{\scriptsize$\pm$0.070} & 0.760{\scriptsize$\pm$0.037} & 0.520{\scriptsize$\pm$0.055} & 0.613{\scriptsize$\pm$0.050} & 0.462{\scriptsize$\pm$0.059} & 0.752{\scriptsize$\pm$0.038} \\
Phi-4-Reasoning & \checkmark & 0.757{\scriptsize$\pm$0.033} & 0.757{\scriptsize$\pm$0.039} & 0.725{\scriptsize$\pm$0.041} & 0.574{\scriptsize$\pm$0.056} & 0.791{\scriptsize$\pm$0.033} & 0.464{\scriptsize$\pm$0.060} & 0.669{\scriptsize$\pm$0.049} & 0.740{\scriptsize$\pm$0.038} & 0.800{\scriptsize$\pm$0.029} \\
\bottomrule
\end{tabular}}
\caption{Property Estimation - Log P ($\rho$)}
\label{property_estimation_logp_spearman_rho}
\end{table}

\begin{table}[H]
\centering
\resizebox{\textwidth}{!}{\begin{tabular}{llccccccccc}
\toprule
Model & Reasoning & Canonical SMILES & Isomeric SMILES & Randomized SMILES & DeepSMILES & IUPAC & SELFIES & MolJSON & CML & InChI \\
\midrule
Qwen3-4B & \checkmark & 0.642{\scriptsize$\pm$0.051} & 0.719{\scriptsize$\pm$0.043} & 0.548{\scriptsize$\pm$0.060} & 0.416{\scriptsize$\pm$0.067} & 0.404{\scriptsize$\pm$0.066} & 0.586{\scriptsize$\pm$0.056} & 0.396{\scriptsize$\pm$0.062} & 0.496{\scriptsize$\pm$0.064} & 0.489{\scriptsize$\pm$0.060} \\
Qwen3-4B & \texttimes & 0.626{\scriptsize$\pm$0.051} & 0.673{\scriptsize$\pm$0.051} & 0.632{\scriptsize$\pm$0.050} & 0.378{\scriptsize$\pm$0.065} & 0.546{\scriptsize$\pm$0.053} & 0.620{\scriptsize$\pm$0.056} & 0.418{\scriptsize$\pm$0.061} & 0.505{\scriptsize$\pm$0.063} & 0.527{\scriptsize$\pm$0.048} \\
Qwen3-30B & \checkmark & 0.914{\scriptsize$\pm$0.036} & 0.959{\scriptsize$\pm$0.008} & 0.861{\scriptsize$\pm$0.046} & 0.820{\scriptsize$\pm$0.047} & 0.954{\scriptsize$\pm$0.008} & 0.830{\scriptsize$\pm$0.054} & 0.972{\scriptsize$\pm$0.007} & 0.975{\scriptsize$\pm$0.005} & 0.973{\scriptsize$\pm$0.006} \\
Qwen3-30B & \texttimes & 0.936{\scriptsize$\pm$0.026} & 0.967{\scriptsize$\pm$0.005} & 0.927{\scriptsize$\pm$0.028} & 0.798{\scriptsize$\pm$0.052} & 0.928{\scriptsize$\pm$0.028} & 0.739{\scriptsize$\pm$0.068} & 0.975{\scriptsize$\pm$0.005} & 0.968{\scriptsize$\pm$0.006} & 0.976{\scriptsize$\pm$0.005} \\
Phi-4 & \checkmark & 0.449{\scriptsize$\pm$0.066} & 0.399{\scriptsize$\pm$0.072} & 0.393{\scriptsize$\pm$0.065} & 0.390{\scriptsize$\pm$0.067} & 0.516{\scriptsize$\pm$0.067} & 0.195{\scriptsize$\pm$0.066} & 0.404{\scriptsize$\pm$0.064} & 0.585{\scriptsize$\pm$0.056} & 0.369{\scriptsize$\pm$0.063} \\
Phi-4-Reasoning & \checkmark & 0.887{\scriptsize$\pm$0.031} & 0.891{\scriptsize$\pm$0.033} & 0.820{\scriptsize$\pm$0.050} & 0.791{\scriptsize$\pm$0.061} & 0.855{\scriptsize$\pm$0.043} & 0.668{\scriptsize$\pm$0.065} & 0.756{\scriptsize$\pm$0.056} & 0.743{\scriptsize$\pm$0.051} & 0.866{\scriptsize$\pm$0.032} \\
\bottomrule
\end{tabular}}
\caption{Property Estimation - TPSA ($\rho$)}
\label{tab:property_estimation_tpsa_spearman_rho}
\end{table}

\begin{table}[H]
\centering
\resizebox{\textwidth}{!}{\begin{tabular}{llccccccccc}
\toprule
Model & Reasoning & Canonical SMILES & Isomeric SMILES & Randomized SMILES & DeepSMILES & IUPAC & SELFIES & MolJSON & CML & InChI \\
\midrule
Qwen3-4B & \checkmark & 0.300{\scriptsize$\pm$0.033} & 0.315{\scriptsize$\pm$0.034} & 0.250{\scriptsize$\pm$0.031} & 0.185{\scriptsize$\pm$0.027} & 0.585{\scriptsize$\pm$0.035} & 0.165{\scriptsize$\pm$0.026} & 0.520{\scriptsize$\pm$0.035} & 0.785{\scriptsize$\pm$0.030} & 0.230{\scriptsize$\pm$0.030} \\
Qwen3-4B & \texttimes & 0.290{\scriptsize$\pm$0.032} & 0.285{\scriptsize$\pm$0.033} & 0.310{\scriptsize$\pm$0.033} & 0.205{\scriptsize$\pm$0.029} & 0.550{\scriptsize$\pm$0.034} & 0.140{\scriptsize$\pm$0.023} & 0.505{\scriptsize$\pm$0.036} & 0.775{\scriptsize$\pm$0.030} & 0.230{\scriptsize$\pm$0.030} \\
Qwen3-30B & \checkmark & 0.435{\scriptsize$\pm$0.036} & 0.495{\scriptsize$\pm$0.034} & 0.475{\scriptsize$\pm$0.036} & 0.265{\scriptsize$\pm$0.032} & 0.615{\scriptsize$\pm$0.034} & 0.130{\scriptsize$\pm$0.023} & 0.620{\scriptsize$\pm$0.035} & 0.760{\scriptsize$\pm$0.031} & 0.380{\scriptsize$\pm$0.035} \\
Qwen3-30B & \texttimes & 0.450{\scriptsize$\pm$0.036} & 0.445{\scriptsize$\pm$0.035} & 0.410{\scriptsize$\pm$0.036} & 0.310{\scriptsize$\pm$0.031} & 0.590{\scriptsize$\pm$0.034} & 0.105{\scriptsize$\pm$0.022} & 0.685{\scriptsize$\pm$0.033} & 0.775{\scriptsize$\pm$0.030} & 0.410{\scriptsize$\pm$0.035} \\
Phi-4 & \checkmark & 0.340{\scriptsize$\pm$0.035} & 0.335{\scriptsize$\pm$0.034} & 0.286{\scriptsize$\pm$0.033} & 0.190{\scriptsize$\pm$0.028} & 0.515{\scriptsize$\pm$0.036} & 0.215{\scriptsize$\pm$0.029} & 0.120{\scriptsize$\pm$0.022} & 0.730{\scriptsize$\pm$0.032} & 0.215{\scriptsize$\pm$0.029} \\
Phi-4-Reasoning & \checkmark & 0.350{\scriptsize$\pm$0.034} & 0.365{\scriptsize$\pm$0.034} & 0.330{\scriptsize$\pm$0.034} & 0.235{\scriptsize$\pm$0.030} & 0.583{\scriptsize$\pm$0.037} & 0.111{\scriptsize$\pm$0.022} & 0.410{\scriptsize$\pm$0.034} & 0.780{\scriptsize$\pm$0.030} & 0.410{\scriptsize$\pm$0.035} \\
\bottomrule
\end{tabular}}
\caption{Property Estimation - HBD (Exact Match)}
\label{tab:property_estimation_hbd_exact_match}
\end{table}

\begin{table}[H]
\centering
\resizebox{\textwidth}{!}{\begin{tabular}{llccccccccc}
\toprule
Model & Reasoning & Canonical SMILES & Isomeric SMILES & Randomized SMILES & DeepSMILES & IUPAC & SELFIES & MolJSON & CML & InChI \\
\midrule
Qwen3-4B & \checkmark & 0.385{\scriptsize$\pm$0.034} & 0.395{\scriptsize$\pm$0.034} & 0.280{\scriptsize$\pm$0.032} & 0.325{\scriptsize$\pm$0.032} & 0.365{\scriptsize$\pm$0.034} & 0.210{\scriptsize$\pm$0.029} & 0.330{\scriptsize$\pm$0.033} & 0.485{\scriptsize$\pm$0.034} & 0.465{\scriptsize$\pm$0.035} \\
Qwen3-4B & \texttimes & 0.325{\scriptsize$\pm$0.033} & 0.380{\scriptsize$\pm$0.034} & 0.275{\scriptsize$\pm$0.032} & 0.340{\scriptsize$\pm$0.032} & 0.355{\scriptsize$\pm$0.034} & 0.235{\scriptsize$\pm$0.030} & 0.335{\scriptsize$\pm$0.033} & 0.475{\scriptsize$\pm$0.035} & 0.445{\scriptsize$\pm$0.035} \\
Qwen3-30B & \checkmark & 0.425{\scriptsize$\pm$0.035} & 0.455{\scriptsize$\pm$0.036} & 0.385{\scriptsize$\pm$0.035} & 0.450{\scriptsize$\pm$0.035} & 0.420{\scriptsize$\pm$0.035} & 0.240{\scriptsize$\pm$0.030} & 0.500{\scriptsize$\pm$0.035} & 0.475{\scriptsize$\pm$0.034} & 0.465{\scriptsize$\pm$0.036} \\
Qwen3-30B & \texttimes & 0.450{\scriptsize$\pm$0.036} & 0.420{\scriptsize$\pm$0.035} & 0.410{\scriptsize$\pm$0.035} & 0.455{\scriptsize$\pm$0.035} & 0.430{\scriptsize$\pm$0.035} & 0.230{\scriptsize$\pm$0.030} & 0.515{\scriptsize$\pm$0.034} & 0.460{\scriptsize$\pm$0.034} & 0.485{\scriptsize$\pm$0.035} \\
Phi-4 & \checkmark & 0.255{\scriptsize$\pm$0.031} & 0.210{\scriptsize$\pm$0.029} & 0.260{\scriptsize$\pm$0.031} & 0.165{\scriptsize$\pm$0.026} & 0.270{\scriptsize$\pm$0.031} & 0.100{\scriptsize$\pm$0.022} & 0.440{\scriptsize$\pm$0.033} & 0.390{\scriptsize$\pm$0.034} & 0.365{\scriptsize$\pm$0.034} \\
Phi-4-Reasoning & \checkmark & 0.415{\scriptsize$\pm$0.035} & 0.435{\scriptsize$\pm$0.034} & 0.430{\scriptsize$\pm$0.035} & 0.430{\scriptsize$\pm$0.034} & 0.345{\scriptsize$\pm$0.033} & 0.187{\scriptsize$\pm$0.027} & 0.470{\scriptsize$\pm$0.035} & 0.445{\scriptsize$\pm$0.034} & 0.485{\scriptsize$\pm$0.035} \\
\bottomrule
\end{tabular}}
\caption{Property Estimation - HBA (Exact-Match)}
\label{tab:property_estimation_hba_exact_match}
\end{table}

\begin{table}[H]
\centering
\resizebox{\textwidth}{!}{\begin{tabular}{llccccccccc}
\toprule
Model & Reasoning & Canonical SMILES & Isomeric SMILES & Randomized SMILES & DeepSMILES & IUPAC & SELFIES & MolJSON & CML & InChI \\
\midrule
Qwen3-4B & \checkmark & 0.821{\scriptsize$\pm$0.044} & 0.769{\scriptsize$\pm$0.048} & 0.808{\scriptsize$\pm$0.045} & 0.808{\scriptsize$\pm$0.045} & 0.885{\scriptsize$\pm$0.036} & 0.615{\scriptsize$\pm$0.055} & 0.795{\scriptsize$\pm$0.046} & 0.282{\scriptsize$\pm$0.051} & 0.795{\scriptsize$\pm$0.046} \\
Qwen3-4B & \texttimes & 0.795{\scriptsize$\pm$0.046} & 0.859{\scriptsize$\pm$0.040} & 0.705{\scriptsize$\pm$0.052} & 0.667{\scriptsize$\pm$0.054} & 0.910{\scriptsize$\pm$0.033} & 0.615{\scriptsize$\pm$0.055} & 0.808{\scriptsize$\pm$0.045} & 0.295{\scriptsize$\pm$0.052} & 0.769{\scriptsize$\pm$0.048} \\
Qwen3-30B & \checkmark & 0.885{\scriptsize$\pm$0.036} & 0.910{\scriptsize$\pm$0.033} & 0.833{\scriptsize$\pm$0.042} & 0.872{\scriptsize$\pm$0.038} & 0.910{\scriptsize$\pm$0.033} & 0.692{\scriptsize$\pm$0.053} & 0.795{\scriptsize$\pm$0.046} & 0.244{\scriptsize$\pm$0.049} & 0.897{\scriptsize$\pm$0.035} \\
Qwen3-30B & \texttimes & 0.897{\scriptsize$\pm$0.035} & 0.897{\scriptsize$\pm$0.035} & 0.821{\scriptsize$\pm$0.044} & 0.833{\scriptsize$\pm$0.042} & 0.936{\scriptsize$\pm$0.028} & 0.679{\scriptsize$\pm$0.053} & 0.795{\scriptsize$\pm$0.046} & 0.295{\scriptsize$\pm$0.052} & 0.859{\scriptsize$\pm$0.040} \\
Phi-4 & \checkmark & 0.603{\scriptsize$\pm$0.056} & 0.641{\scriptsize$\pm$0.055} & 0.564{\scriptsize$\pm$0.057} & 0.577{\scriptsize$\pm$0.056} & 0.705{\scriptsize$\pm$0.052} & 0.385{\scriptsize$\pm$0.055} & 0.462{\scriptsize$\pm$0.057} & 0.205{\scriptsize$\pm$0.046} & 0.551{\scriptsize$\pm$0.057} \\
Phi-4-Reasoning & \checkmark & 0.808{\scriptsize$\pm$0.045} & 0.859{\scriptsize$\pm$0.040} & 0.769{\scriptsize$\pm$0.048} & 0.756{\scriptsize$\pm$0.049} & 0.872{\scriptsize$\pm$0.038} & 0.641{\scriptsize$\pm$0.055} & 0.744{\scriptsize$\pm$0.050} & 0.269{\scriptsize$\pm$0.051} & 0.795{\scriptsize$\pm$0.046} \\
\bottomrule
\end{tabular}}
\caption{Molecule Retrieval (accuracy)}
\label{tab:molecule_retrieval_accuracy}
\end{table}

\begin{table}[H]
\centering
\resizebox{\textwidth}{!}{\begin{tabular}{llccccccccc}
\toprule
Model & Reasoning & Canonical SMILES & Isomeric SMILES & Randomized SMILES & DeepSMILES & IUPAC & SELFIES & MolJSON & CML & InChI \\
\midrule
Qwen3-4B & \checkmark & 0.252{\scriptsize$\pm$0.019} & 0.266{\scriptsize$\pm$0.020} & 0.602{\scriptsize$\pm$0.022} & 0.268{\scriptsize$\pm$0.020} & 0.253{\scriptsize$\pm$0.019} & 0.260{\scriptsize$\pm$0.020} & 0.648{\scriptsize$\pm$0.021} & 0.658{\scriptsize$\pm$0.021} & 0.250{\scriptsize$\pm$0.019} \\
Qwen3-4B & \texttimes & 0.262{\scriptsize$\pm$0.020} & 0.262{\scriptsize$\pm$0.020} & 0.610{\scriptsize$\pm$0.022} & 0.276{\scriptsize$\pm$0.020} & 0.251{\scriptsize$\pm$0.019} & 0.276{\scriptsize$\pm$0.020} & 0.638{\scriptsize$\pm$0.022} & 0.660{\scriptsize$\pm$0.021} & 0.252{\scriptsize$\pm$0.019} \\
Qwen3-30B & \checkmark & 0.250{\scriptsize$\pm$0.019} & 0.258{\scriptsize$\pm$0.020} & 0.612{\scriptsize$\pm$0.022} & 0.260{\scriptsize$\pm$0.020} & 0.251{\scriptsize$\pm$0.019} & 0.256{\scriptsize$\pm$0.020} & 0.632{\scriptsize$\pm$0.022} & 0.640{\scriptsize$\pm$0.021} & 0.250{\scriptsize$\pm$0.019} \\
Qwen3-30B & \texttimes & 0.252{\scriptsize$\pm$0.019} & 0.252{\scriptsize$\pm$0.019} & 0.608{\scriptsize$\pm$0.022} & 0.250{\scriptsize$\pm$0.019} & 0.251{\scriptsize$\pm$0.019} & 0.258{\scriptsize$\pm$0.020} & 0.640{\scriptsize$\pm$0.021} & 0.640{\scriptsize$\pm$0.021} & 0.250{\scriptsize$\pm$0.019} \\
Phi-4 & \checkmark & 0.252{\scriptsize$\pm$0.019} & \textbf{0.272{\scriptsize$\pm$0.020}} & 0.462{\scriptsize$\pm$0.022} & 0.272{\scriptsize$\pm$0.020} & 0.248{\scriptsize$\pm$0.019} & 0.290{\scriptsize$\pm$0.020} & 0.620{\scriptsize$\pm$0.022} & 0.604{\scriptsize$\pm$0.022} & 0.254{\scriptsize$\pm$0.019} \\
\bottomrule
\end{tabular}}
\caption{Isomer Discrimination (accuracy)}
\label{tab:isomer_discrimination_accuracy}
\end{table}

\begin{table}[H]
\centering
\resizebox{\textwidth}{!}{\begin{tabular}{llccccccccc}
\toprule
Model & Reasoning & Canonical SMILES & Isomeric SMILES & Randomized SMILES & DeepSMILES & IUPAC & SELFIES & MolJSON & CML & InChI \\
\midrule
Qwen3-4B & \checkmark & 0.624{\scriptsize$\pm$0.031} & 0.584{\scriptsize$\pm$0.031} & 0.652{\scriptsize$\pm$0.030} & 0.016{\scriptsize$\pm$0.008} & 0.600{\scriptsize$\pm$0.031} & \textbf{0.884{\scriptsize$\pm$0.020}} & 0.708{\scriptsize$\pm$0.029} & 0.336{\scriptsize$\pm$0.030} & 0.004{\scriptsize$\pm$0.004} \\
Qwen3-4B & \texttimes & 0.592{\scriptsize$\pm$0.031} & 0.552{\scriptsize$\pm$0.032} & 0.596{\scriptsize$\pm$0.031} & 0.016{\scriptsize$\pm$0.008} & 0.588{\scriptsize$\pm$0.031} & 0.876{\scriptsize$\pm$0.021} & 0.748{\scriptsize$\pm$0.028} & 0.344{\scriptsize$\pm$0.030} & 0.004{\scriptsize$\pm$0.004} \\
Qwen3-30B & \checkmark & 0.712{\scriptsize$\pm$0.029} & 0.692{\scriptsize$\pm$0.029} & 0.696{\scriptsize$\pm$0.029} & 0.008{\scriptsize$\pm$0.006} & 0.644{\scriptsize$\pm$0.030} & 0.532{\scriptsize$\pm$0.032} & 0.560{\scriptsize$\pm$0.031} & 0.288{\scriptsize$\pm$0.029} & 0.032{\scriptsize$\pm$0.011} \\
Qwen3-30B & \texttimes & 0.700{\scriptsize$\pm$0.029} & 0.624{\scriptsize$\pm$0.031} & 0.700{\scriptsize$\pm$0.029} & 0.000 & 0.660{\scriptsize$\pm$0.030} & 0.512{\scriptsize$\pm$0.032} & 0.612{\scriptsize$\pm$0.031} & 0.276{\scriptsize$\pm$0.028} & 0.036{\scriptsize$\pm$0.012} \\
Phi-4 & \checkmark & 0.712{\scriptsize$\pm$0.029} & 0.712{\scriptsize$\pm$0.029} & 0.716{\scriptsize$\pm$0.029} & 0.008{\scriptsize$\pm$0.006} & 0.484{\scriptsize$\pm$0.032} & 0.080{\scriptsize$\pm$0.017} & 0.056{\scriptsize$\pm$0.015} & 0.004{\scriptsize$\pm$0.004} & 0.020{\scriptsize$\pm$0.009} \\
Phi-4-Reasoning & \checkmark & 0.324{\scriptsize$\pm$0.030} & 0.240{\scriptsize$\pm$0.027} & 0.296{\scriptsize$\pm$0.029} & 0.020{\scriptsize$\pm$0.009} & 0.452{\scriptsize$\pm$0.032} & 0.380{\scriptsize$\pm$0.031} & 0.476{\scriptsize$\pm$0.032} & 0.112{\scriptsize$\pm$0.020} & 0.016{\scriptsize$\pm$0.008} \\
\bottomrule
\end{tabular}}
\caption{Caption-to-Molecule (Validity Rate)}
\label{tab:caption_to_molecule_validity_rate}
\end{table}

\begin{table}[H]
\centering
\resizebox{\textwidth}{!}{\begin{tabular}{llccccccccc}
\toprule
Model & Reasoning & Canonical SMILES & Isomeric SMILES & Randomized SMILES & DeepSMILES & IUPAC & SELFIES & MolJSON & CML & InChI \\
\midrule
Qwen3-4B & \checkmark & 0.052{\scriptsize$\pm$0.014} & 0.036{\scriptsize$\pm$0.012} & 0.044{\scriptsize$\pm$0.013} & 0.008{\scriptsize$\pm$0.006} & 0.292{\scriptsize$\pm$0.029} & 0.000 & 0.040{\scriptsize$\pm$0.012} & 0.000 & 0.000 \\
Qwen3-4B & \texttimes & 0.052{\scriptsize$\pm$0.014} & 0.040{\scriptsize$\pm$0.012} & 0.036{\scriptsize$\pm$0.012} & 0.008{\scriptsize$\pm$0.006} & 0.296{\scriptsize$\pm$0.029} & 0.000 & 0.052{\scriptsize$\pm$0.014} & 0.000 & 0.000 \\
Qwen3-30B & \checkmark & 0.056{\scriptsize$\pm$0.015} & 0.080{\scriptsize$\pm$0.017} & 0.056{\scriptsize$\pm$0.015} & 0.008{\scriptsize$\pm$0.006} & \textbf{0.348{\scriptsize$\pm$0.030}} & 0.008{\scriptsize$\pm$0.006} & 0.052{\scriptsize$\pm$0.014} & 0.012{\scriptsize$\pm$0.007} & 0.008{\scriptsize$\pm$0.006} \\
Qwen3-30B & \texttimes & 0.080{\scriptsize$\pm$0.017} & 0.056{\scriptsize$\pm$0.015} & 0.076{\scriptsize$\pm$0.017} & 0.000 & 0.340{\scriptsize$\pm$0.030} & 0.000 & 0.068{\scriptsize$\pm$0.016} & 0.004{\scriptsize$\pm$0.004} & 0.008{\scriptsize$\pm$0.006} \\
Phi-4 & \checkmark & 0.016{\scriptsize$\pm$0.008} & 0.020{\scriptsize$\pm$0.009} & 0.016{\scriptsize$\pm$0.008} & 0.004{\scriptsize$\pm$0.004} & 0.184{\scriptsize$\pm$0.025} & 0.000 & 0.012{\scriptsize$\pm$0.007} & 0.000 & 0.012{\scriptsize$\pm$0.007} \\
Phi-4-Reasoning & \checkmark & 0.048{\scriptsize$\pm$0.014} & 0.032{\scriptsize$\pm$0.011} & 0.044{\scriptsize$\pm$0.013} & 0.000 & 0.248{\scriptsize$\pm$0.027} & 0.004{\scriptsize$\pm$0.004} & 0.056{\scriptsize$\pm$0.015} & 0.016{\scriptsize$\pm$0.008} & 0.004{\scriptsize$\pm$0.004} \\
\bottomrule
\end{tabular}}
\caption{Caption-to-Molecule (Exact Match)}
\label{tab:caption_to_molecule_exact_match}
\end{table}

\begin{table}[H]
\centering
\resizebox{\textwidth}{!}{\begin{tabular}{llccccccccc}
\toprule
Model & Reasoning & Canonical SMILES & Isomeric SMILES & Randomized SMILES & DeepSMILES & IUPAC & SELFIES & MolJSON & CML & InChI \\
\midrule
Qwen3-4B & \checkmark & 0.203{\scriptsize$\pm$0.017} & 0.193{\scriptsize$\pm$0.017} & 0.212{\scriptsize$\pm$0.017} & 0.013{\scriptsize$\pm$0.007} & 0.439{\scriptsize$\pm$0.027} & 0.000 & 0.154{\scriptsize$\pm$0.017} & 0.000{\scriptsize$\pm$0.000} & 0.000{\scriptsize$\pm$0.000} \\
Qwen3-4B & \texttimes & 0.200{\scriptsize$\pm$0.018} & 0.178{\scriptsize$\pm$0.016} & 0.192{\scriptsize$\pm$0.017} & 0.013{\scriptsize$\pm$0.007} & 0.428{\scriptsize$\pm$0.027} & 0.000 & 0.166{\scriptsize$\pm$0.017} & 0.000{\scriptsize$\pm$0.000} & 0.000{\scriptsize$\pm$0.000} \\
Qwen3-30B & \checkmark & 0.275{\scriptsize$\pm$0.018} & 0.269{\scriptsize$\pm$0.019} & 0.260{\scriptsize$\pm$0.018} & 0.008{\scriptsize$\pm$0.006} & 0.494{\scriptsize$\pm$0.027} & 0.047{\scriptsize$\pm$0.007} & 0.180{\scriptsize$\pm$0.020} & 0.029{\scriptsize$\pm$0.008} & 0.014{\scriptsize$\pm$0.006} \\
Qwen3-30B & \texttimes & 0.269{\scriptsize$\pm$0.019} & 0.233{\scriptsize$\pm$0.018} & 0.286{\scriptsize$\pm$0.019} & 0.000 & 0.493{\scriptsize$\pm$0.027} & 0.037{\scriptsize$\pm$0.005} & 0.213{\scriptsize$\pm$0.021} & 0.013{\scriptsize$\pm$0.005} & 0.014{\scriptsize$\pm$0.006} \\
Phi-4 & \checkmark & 0.208{\scriptsize$\pm$0.013} & 0.198{\scriptsize$\pm$0.013} & 0.200{\scriptsize$\pm$0.013} & 0.005{\scriptsize$\pm$0.004} & 0.315{\scriptsize$\pm$0.025} & 0.004{\scriptsize$\pm$0.002} & 0.018{\scriptsize$\pm$0.007} & 0.000 & 0.017{\scriptsize$\pm$0.008} \\
Phi-4-Reasoning & \checkmark & 0.146{\scriptsize$\pm$0.017} & 0.097{\scriptsize$\pm$0.015} & 0.126{\scriptsize$\pm$0.016} & 0.006{\scriptsize$\pm$0.004} & 0.341{\scriptsize$\pm$0.027} & 0.021{\scriptsize$\pm$0.005} & 0.153{\scriptsize$\pm$0.018} & 0.031{\scriptsize$\pm$0.009} & 0.010{\scriptsize$\pm$0.005} \\
\bottomrule
\end{tabular}}
\caption{Caption-to-Molecule (Morgan Tanimoto Similarity)}
\label{tab:caption_to_molecule_mean_tanimoto}
\end{table}

\begin{table}[H]
\centering
\resizebox{\textwidth}{!}{\begin{tabular}{llccccccccc}
\toprule
Model & Reasoning & Canonical SMILES & Isomeric SMILES & Randomized SMILES & DeepSMILES & IUPAC & SELFIES & MolJSON & CML & InChI \\
\midrule
Qwen3-4B & \checkmark & 0.389{\scriptsize$\pm$0.022} & 0.371{\scriptsize$\pm$0.022} & 0.410{\scriptsize$\pm$0.022} & 0.014{\scriptsize$\pm$0.007} & 0.546{\scriptsize$\pm$0.029} & 0.002{\scriptsize$\pm$0.001} & 0.331{\scriptsize$\pm$0.020} & 0.013{\scriptsize$\pm$0.003} & 0.001{\scriptsize$\pm$0.001} \\
Qwen3-4B & \texttimes & 0.383{\scriptsize$\pm$0.023} & 0.363{\scriptsize$\pm$0.023} & 0.387{\scriptsize$\pm$0.023} & 0.015{\scriptsize$\pm$0.007} & 0.537{\scriptsize$\pm$0.029} & 0.002{\scriptsize$\pm$0.001} & 0.350{\scriptsize$\pm$0.020} & 0.021{\scriptsize$\pm$0.004} & 0.001{\scriptsize$\pm$0.001} \\
Qwen3-30B & \checkmark & 0.518{\scriptsize$\pm$0.023} & 0.503{\scriptsize$\pm$0.023} & 0.497{\scriptsize$\pm$0.023} & 0.008{\scriptsize$\pm$0.006} & 0.603{\scriptsize$\pm$0.029} & 0.105{\scriptsize$\pm$0.011} & 0.320{\scriptsize$\pm$0.023} & 0.078{\scriptsize$\pm$0.012} & 0.020{\scriptsize$\pm$0.008} \\
Qwen3-30B & \texttimes & 0.514{\scriptsize$\pm$0.023} & 0.456{\scriptsize$\pm$0.024} & 0.518{\scriptsize$\pm$0.024} & 0.000 & 0.605{\scriptsize$\pm$0.028} & 0.101{\scriptsize$\pm$0.010} & \textbf{0.362{\scriptsize$\pm$0.024}} & 0.048{\scriptsize$\pm$0.008} & 0.025{\scriptsize$\pm$0.009} \\
Phi-4 & \checkmark & 0.484{\scriptsize$\pm$0.021} & 0.475{\scriptsize$\pm$0.021} & 0.470{\scriptsize$\pm$0.021} & 0.006{\scriptsize$\pm$0.004} & 0.421{\scriptsize$\pm$0.028} & 0.010{\scriptsize$\pm$0.004} & 0.033{\scriptsize$\pm$0.009} & 0.000 & 0.019{\scriptsize$\pm$0.008} \\
Phi-4-Reasoning & \checkmark & 0.239{\scriptsize$\pm$0.023} & 0.165{\scriptsize$\pm$0.020} & 0.220{\scriptsize$\pm$0.022} & 0.008{\scriptsize$\pm$0.005} & 0.415{\scriptsize$\pm$0.030} & 0.045{\scriptsize$\pm$0.007} & 0.285{\scriptsize$\pm$0.023} & 0.058{\scriptsize$\pm$0.012} & 0.013{\scriptsize$\pm$0.007} \\
\bottomrule
\end{tabular}}
\caption{Caption-to-Molecule (MACCS Tanimoto Similarity)}
\label{tab:caption_to_molecule_maccs_tanimoto}
\end{table}

\begin{table}[H]
\centering
\resizebox{\textwidth}{!}{\begin{tabular}{llccccccccc}
\toprule
Model & Reasoning & Canonical SMILES & Isomeric SMILES & Randomized SMILES & DeepSMILES & IUPAC & SELFIES & MolJSON & CML & InChI \\
\midrule
Qwen3-4B & \checkmark & 0.257{\scriptsize$\pm$0.019} & 0.248{\scriptsize$\pm$0.019} & 0.259{\scriptsize$\pm$0.019} & 0.014{\scriptsize$\pm$0.007} & 0.475{\scriptsize$\pm$0.028} & 0.004{\scriptsize$\pm$0.004} & 0.160{\scriptsize$\pm$0.018} & 0.000 & 0.001{\scriptsize$\pm$0.001} \\
Qwen3-4B & \texttimes & 0.248{\scriptsize$\pm$0.019} & 0.233{\scriptsize$\pm$0.019} & 0.242{\scriptsize$\pm$0.019} & 0.014{\scriptsize$\pm$0.007} & 0.458{\scriptsize$\pm$0.028} & 0.004{\scriptsize$\pm$0.004} & 0.182{\scriptsize$\pm$0.019} & 0.000 & 0.001{\scriptsize$\pm$0.001} \\
Qwen3-30B & \checkmark & 0.356{\scriptsize$\pm$0.020} & 0.347{\scriptsize$\pm$0.020} & 0.330{\scriptsize$\pm$0.019} & 0.008{\scriptsize$\pm$0.006} & \textbf{0.532{\scriptsize$\pm$0.028}} & 0.041{\scriptsize$\pm$0.008} & 0.195{\scriptsize$\pm$0.021} & 0.040{\scriptsize$\pm$0.010} & 0.019{\scriptsize$\pm$0.008} \\
Qwen3-30B & \texttimes & 0.350{\scriptsize$\pm$0.020} & 0.309{\scriptsize$\pm$0.020} & 0.365{\scriptsize$\pm$0.021} & 0.000 & 0.527{\scriptsize$\pm$0.028} & 0.034{\scriptsize$\pm$0.006} & 0.226{\scriptsize$\pm$0.022} & 0.021{\scriptsize$\pm$0.007} & 0.025{\scriptsize$\pm$0.009} \\
Phi-4 & \checkmark & 0.307{\scriptsize$\pm$0.017} & 0.289{\scriptsize$\pm$0.016} & 0.286{\scriptsize$\pm$0.016} & 0.006{\scriptsize$\pm$0.004} & 0.357{\scriptsize$\pm$0.027} & 0.003{\scriptsize$\pm$0.001} & 0.020{\scriptsize$\pm$0.007} & 0.000{\scriptsize$\pm$0.000} & 0.017{\scriptsize$\pm$0.008} \\
Phi-4-Reasoning & \checkmark & 0.173{\scriptsize$\pm$0.019} & 0.112{\scriptsize$\pm$0.016} & 0.158{\scriptsize$\pm$0.019} & 0.007{\scriptsize$\pm$0.005} & 0.362{\scriptsize$\pm$0.028} & 0.015{\scriptsize$\pm$0.005} & 0.180{\scriptsize$\pm$0.019} & 0.034{\scriptsize$\pm$0.010} & 0.012{\scriptsize$\pm$0.006} \\
\bottomrule
\end{tabular}}
\caption{Caption-to-Molecule (RDKit Tanimoto Similarity)}
\label{tab:caption_to_molecule_rdkit_tanimoto}
\end{table}

\begin{table}[H]
\centering
\resizebox{\textwidth}{!}{\begin{tabular}{llccccccccc}
\toprule
Model & Reasoning & Canonical SMILES & Isomeric SMILES & Randomized SMILES & DeepSMILES & IUPAC & SELFIES & MolJSON & CML & InChI \\
\midrule
Qwen3-4B & \checkmark & 12.069 & 12.286 & 11.798 & 51.767 & 6.248 & 41.612 & 19.528 & 38.580 & N/A \\
Qwen3-4B & \texttimes & 12.454 & 12.186 & 12.784 & 50.127 & 6.969 & 41.610 & 18.420 & 36.003 & N/A \\
Qwen3-30B & \checkmark & 8.141 & 8.058 & 8.657 & 56.978 & \textbf{5.227} & 28.994 & 17.361 & 23.819 & 35.137 \\
Qwen3-30B & \texttimes & 8.650 & 8.467 & 8.227 & N/A & 5.883 & 29.035 & 16.145 & 26.109 & 32.779 \\
Phi-4 & \checkmark & 8.466 & 9.017 & 8.563 & 61.488 & 8.496 & 33.466 & 29.659 & N/A & 42.122 \\
Phi-4-Reasoning & \checkmark & 13.159 & 15.857 & 14.433 & 39.255 & 8.074 & 29.703 & 19.044 & 26.084 & 56.278 \\
Phi-4-Reasoning-Plus & \checkmark & 12.900 & 14.178 & 13.588 & 37.846 & 8.616 & 30.986 & 16.793 & 27.363 & 46.232 \\
Qwen2.5-14B & \texttimes & 12.590 & 13.450 & 13.624 & 39.922 & 9.832 & 38.124 & N/A & 32.164 & 36.180 \\
ChemDFM-v2.0-14B & \texttimes & 5.437 & 5.736 & 10.826 & 49.030 & 17.443 & 41.865 & 26.946 & N/A & N/A \\
ChemDFM-R-14B & \checkmark & \textbf{3.544} & 10.992 & \textbf{3.666} & 43.437 & 21.877 & 41.883 & 23.124 & N/A & N/A \\
Mistral-Small-24B-Instruct-2501 & \texttimes & 10.729 & 9.950 & 10.412 & 45.993 & 7.762 & 32.571 & 39.151 & 27.853 & 35.916 \\
Ether0-24B & \checkmark & N/A & N/A & 39.400 & N/A & N/A & 41.344 & N/A & N/A & N/A \\
OLMo-3.1-32B-Instruct & \texttimes & 17.039 & 18.432 & 17.412 & 38.455 & 11.165 & 34.303 & 23.693 & N/A & N/A \\
OLMo-3.1-32B-Think & \checkmark & 15.712 & 15.881 & 15.173 & 41.208 & 6.909 & 35.765 & 18.712 & N/A & N/A \\
GPT-5.4-Mini & \checkmark & 4.077 & \textbf{3.969} & 3.994 & \textbf{18.590} & 10.428 & \textbf{21.880} & \textbf{12.969} & \textbf{14.820} & \textbf{27.641} \\
Claude-Haiku-4.5 & \checkmark & 5.277 & 5.460 & 5.901 & 40.766 & 5.527 & 21.898 & 15.165 & 15.280 & 33.615 \\
\bottomrule
\end{tabular}}
\caption{Caption-to-Molecule (FCD) (For some models/representations, FCD is not computable due to the absence of valid generations so marked as N/A.)}
\label{tab:caption_to_molecule_fcd}
\end{table}

\begin{table}[H]
\centering
\resizebox{\textwidth}{!}{\begin{tabular}{llccccccccc}
\toprule
Model & Reasoning & Canonical SMILES & Isomeric SMILES & Randomized SMILES & DeepSMILES & IUPAC & SELFIES & MolJSON & CML & InChI \\
\midrule
Qwen3-4B & \checkmark & 0.808{\scriptsize$\pm$0.025} & 0.792{\scriptsize$\pm$0.026} & 0.668{\scriptsize$\pm$0.030} & 0.744{\scriptsize$\pm$0.028} & 0.498{\scriptsize$\pm$0.032} & 0.500{\scriptsize$\pm$0.032} & 0.796{\scriptsize$\pm$0.026} & 0.808{\scriptsize$\pm$0.025} & 0.684{\scriptsize$\pm$0.029} \\
Qwen3-4B & \texttimes & 0.796{\scriptsize$\pm$0.026} & 0.776{\scriptsize$\pm$0.026} & 0.640{\scriptsize$\pm$0.030} & 0.724{\scriptsize$\pm$0.028} & 0.498{\scriptsize$\pm$0.032} & 0.504{\scriptsize$\pm$0.032} & 0.780{\scriptsize$\pm$0.026} & 0.824{\scriptsize$\pm$0.024} & 0.704{\scriptsize$\pm$0.029} \\
Qwen3-30B & \checkmark & 0.832{\scriptsize$\pm$0.024} & 0.780{\scriptsize$\pm$0.026} & 0.740{\scriptsize$\pm$0.028} & 0.828{\scriptsize$\pm$0.024} & 0.498{\scriptsize$\pm$0.032} & 0.488{\scriptsize$\pm$0.032} & 0.844{\scriptsize$\pm$0.023} & 0.836{\scriptsize$\pm$0.023} & 0.804{\scriptsize$\pm$0.025} \\
Qwen3-30B & \texttimes & 0.812{\scriptsize$\pm$0.025} & 0.820{\scriptsize$\pm$0.024} & 0.708{\scriptsize$\pm$0.029} & 0.864{\scriptsize$\pm$0.022} & 0.498{\scriptsize$\pm$0.032} & 0.492{\scriptsize$\pm$0.032} & 0.864{\scriptsize$\pm$0.022} & 0.848{\scriptsize$\pm$0.023} & 0.824{\scriptsize$\pm$0.024} \\
Phi-4 & \checkmark & 0.872{\scriptsize$\pm$0.021} & 0.864{\scriptsize$\pm$0.022} & 0.820{\scriptsize$\pm$0.024} & 0.856{\scriptsize$\pm$0.022} & 0.498{\scriptsize$\pm$0.032} & 0.440{\scriptsize$\pm$0.031} & \textbf{0.948{\scriptsize$\pm$0.014}} & 0.888{\scriptsize$\pm$0.020} & 0.912{\scriptsize$\pm$0.018} \\
\bottomrule
\end{tabular}}
\caption{Tautomer Recognition (accuracy)}
\label{tab:tautomer_recognition_accuracy}
\end{table}

\begin{table}[H]
\centering
\resizebox{\textwidth}{!}{\begin{tabular}{llccccccccc}
\toprule
Model & Reasoning & Canonical SMILES & Isomeric SMILES & Randomized SMILES & DeepSMILES & IUPAC & SELFIES & MolJSON & CML & InChI \\
\midrule
Qwen3-4B & \checkmark & 0.990{\scriptsize$\pm$0.007} & 0.990{\scriptsize$\pm$0.007} & 0.824{\scriptsize$\pm$0.026} & 0.971{\scriptsize$\pm$0.012} & 0.495{\scriptsize$\pm$0.035} & 0.500{\scriptsize$\pm$0.035} & 0.957{\scriptsize$\pm$0.014} & 0.881{\scriptsize$\pm$0.022} & 0.843{\scriptsize$\pm$0.025} \\
Qwen3-4B & \texttimes & \textbf{0.995{\scriptsize$\pm$0.005}} & 0.981{\scriptsize$\pm$0.009} & 0.819{\scriptsize$\pm$0.027} & 0.976{\scriptsize$\pm$0.011} & 0.495{\scriptsize$\pm$0.035} & 0.500{\scriptsize$\pm$0.035} & 0.967{\scriptsize$\pm$0.012} & 0.929{\scriptsize$\pm$0.018} & 0.848{\scriptsize$\pm$0.025} \\
Qwen3-30B & \checkmark & 0.986{\scriptsize$\pm$0.008} & \textbf{0.995{\scriptsize$\pm$0.005}} & 0.810{\scriptsize$\pm$0.027} & 0.967{\scriptsize$\pm$0.012} & 0.548{\scriptsize$\pm$0.035} & 0.510{\scriptsize$\pm$0.035} & 0.990{\scriptsize$\pm$0.007} & \textbf{0.990{\scriptsize$\pm$0.007}} & 0.781{\scriptsize$\pm$0.029} \\
Qwen3-30B & \texttimes & 0.976{\scriptsize$\pm$0.011} & 0.986{\scriptsize$\pm$0.008} & 0.833{\scriptsize$\pm$0.026} & 0.981{\scriptsize$\pm$0.009} & 0.524{\scriptsize$\pm$0.035} & 0.510{\scriptsize$\pm$0.035} & 0.990{\scriptsize$\pm$0.007} & 0.986{\scriptsize$\pm$0.008} & 0.776{\scriptsize$\pm$0.029} \\
Phi-4 & \checkmark & 0.981{\scriptsize$\pm$0.009} & 0.981{\scriptsize$\pm$0.009} & 0.895{\scriptsize$\pm$0.021} & 0.943{\scriptsize$\pm$0.016} & 0.514{\scriptsize$\pm$0.035} & 0.529{\scriptsize$\pm$0.035} & 0.981{\scriptsize$\pm$0.009} & 0.724{\scriptsize$\pm$0.031} & \textbf{1.000} \\
Phi-4-Reasoning & \checkmark & 0.986{\scriptsize$\pm$0.008} & 0.981{\scriptsize$\pm$0.009} & 0.848{\scriptsize$\pm$0.025} & 0.938{\scriptsize$\pm$0.017} & 0.500{\scriptsize$\pm$0.035} & 0.505{\scriptsize$\pm$0.035} & 0.990{\scriptsize$\pm$0.007} & 0.952{\scriptsize$\pm$0.015} & \textbf{1.000} \\
\bottomrule
\end{tabular}}
\caption{Protonation State Recognition (accuracy)}
\label{tab:protonation_state_recognition_accuracy}
\end{table}

\clearpage

\section{CKA similarity between Qwen2.5 and ChemDFMs across representations}
\label{cka}

\begin{figure}[H]
  \centering

  \begin{subfigure}[t]{0.30\textwidth}
    \centering
    \includegraphics[width=\textwidth]{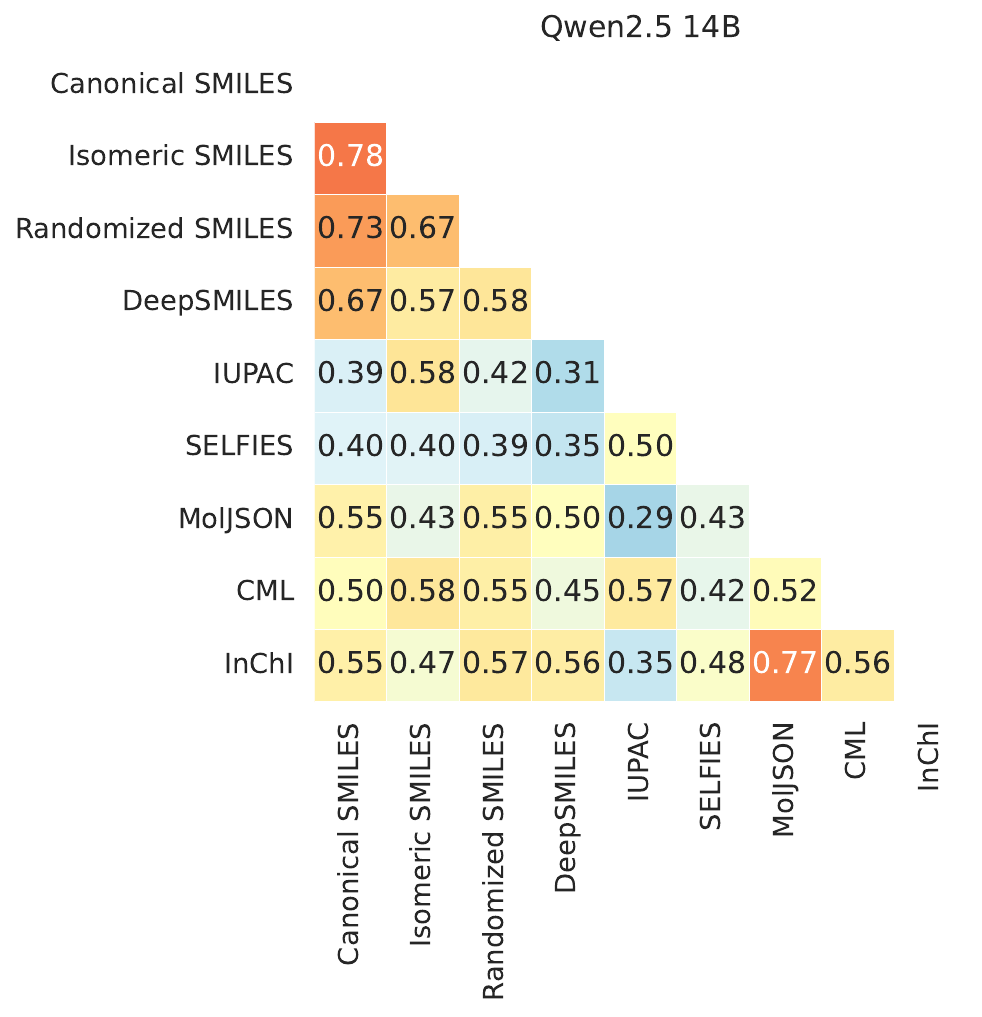}
    \caption{Qwen2.5-14B (base)}
    \label{fig:cka-qwen25}
  \end{subfigure}
  \hfill
  \begin{subfigure}[t]{0.30\textwidth}
    \centering
    \includegraphics[width=\textwidth]{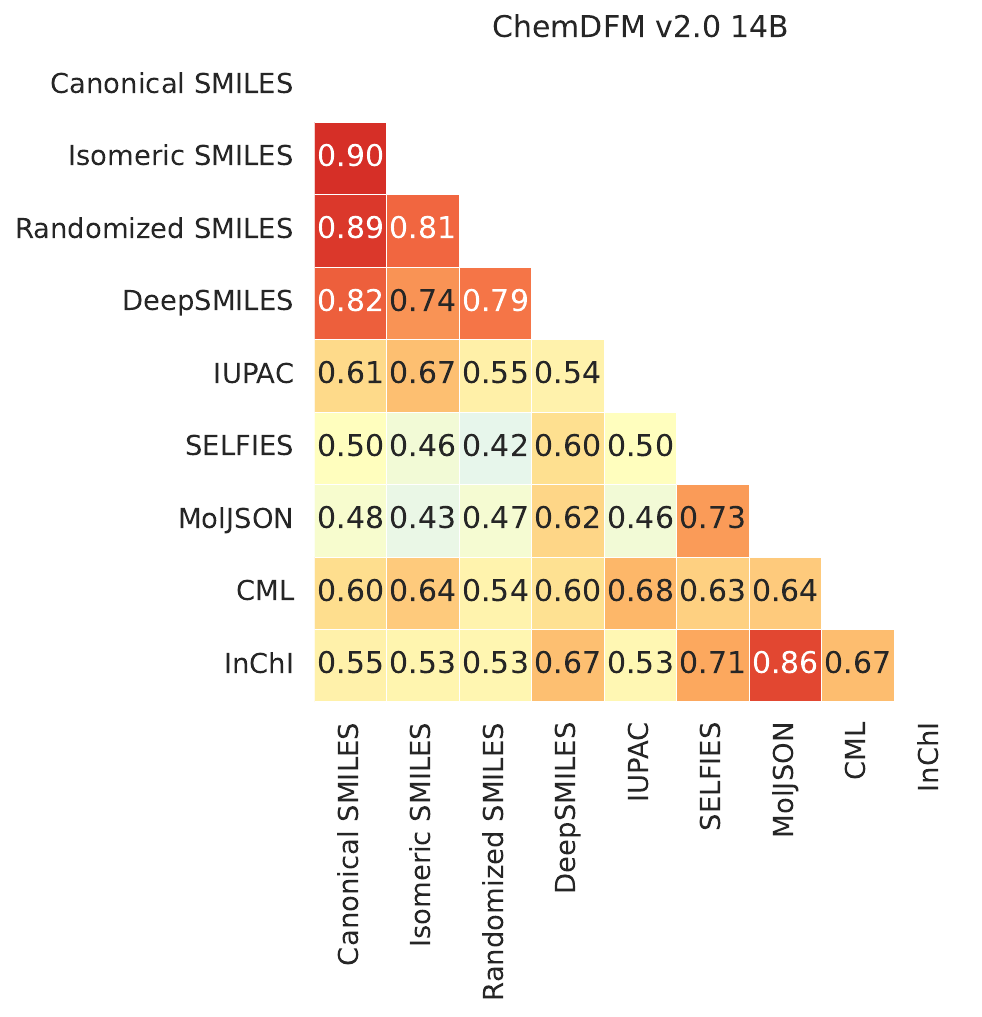}
    \caption{ChemDFM-v2.0-14B}
    \label{fig:cka-chemdfm-v2}
  \end{subfigure}
  \hfill
  \begin{subfigure}[t]{0.30\textwidth}
    \centering
    \includegraphics[width=\textwidth]{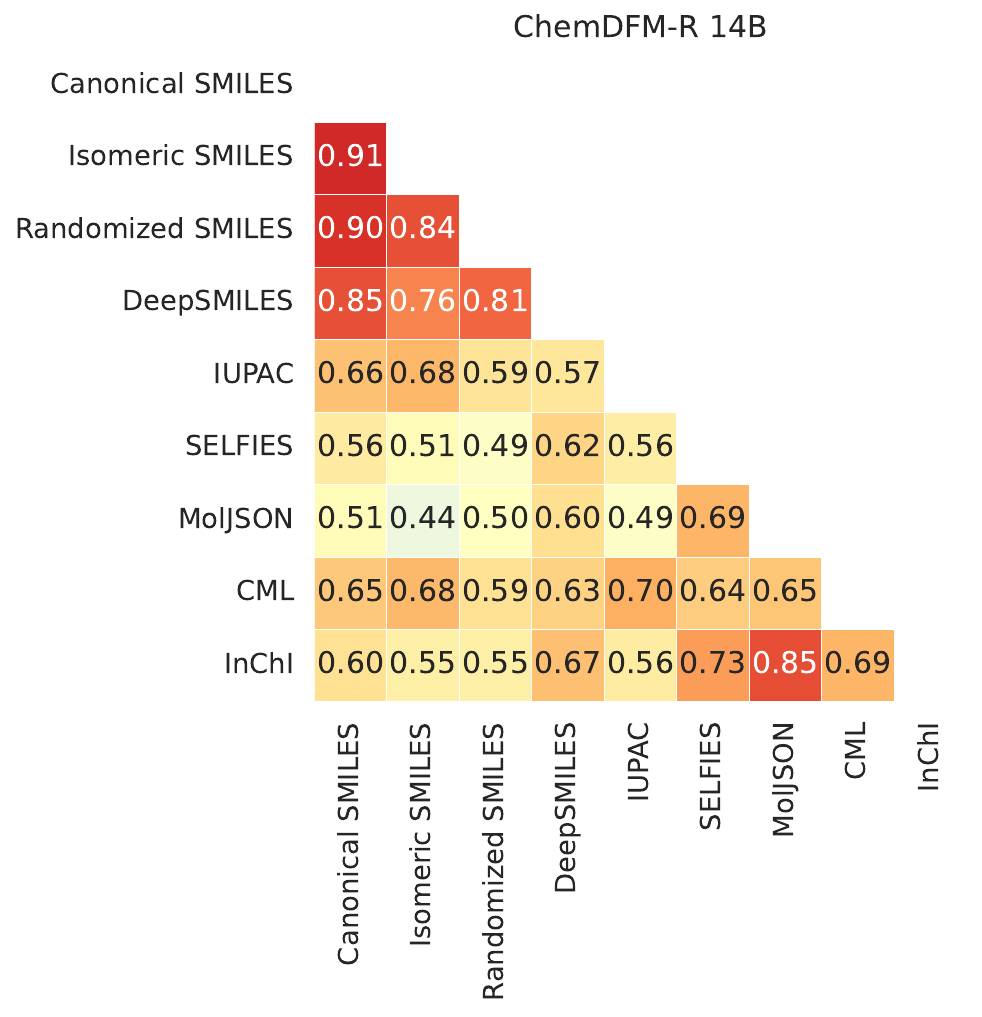}
    \caption{ChemDFM-R-14B}
    \label{fig:cka-chemdfm-r}
  \end{subfigure}
  \hfill
  \begin{subfigure}[t]{0.08\textwidth}
    \centering
    \includegraphics[width=\textwidth]{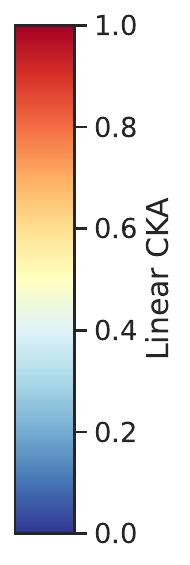}
  \end{subfigure}
  \caption{Linear CKA similarity between representation pairs of chemistry-specialized models, ChemDFM-v2.0 and ChemDFM-R, and their base model Qwen2.5-14B. Higher values (darker red) indicate the model encodes two representations more similarly.}
  \label{fig:cka-comparison}
\end{figure}

\section{t-SNE plots of molecular text representations across different LLMs}
\label{appendix_tsne}

\begin{figure}[H]
  \centering
  \includegraphics[width=\columnwidth]{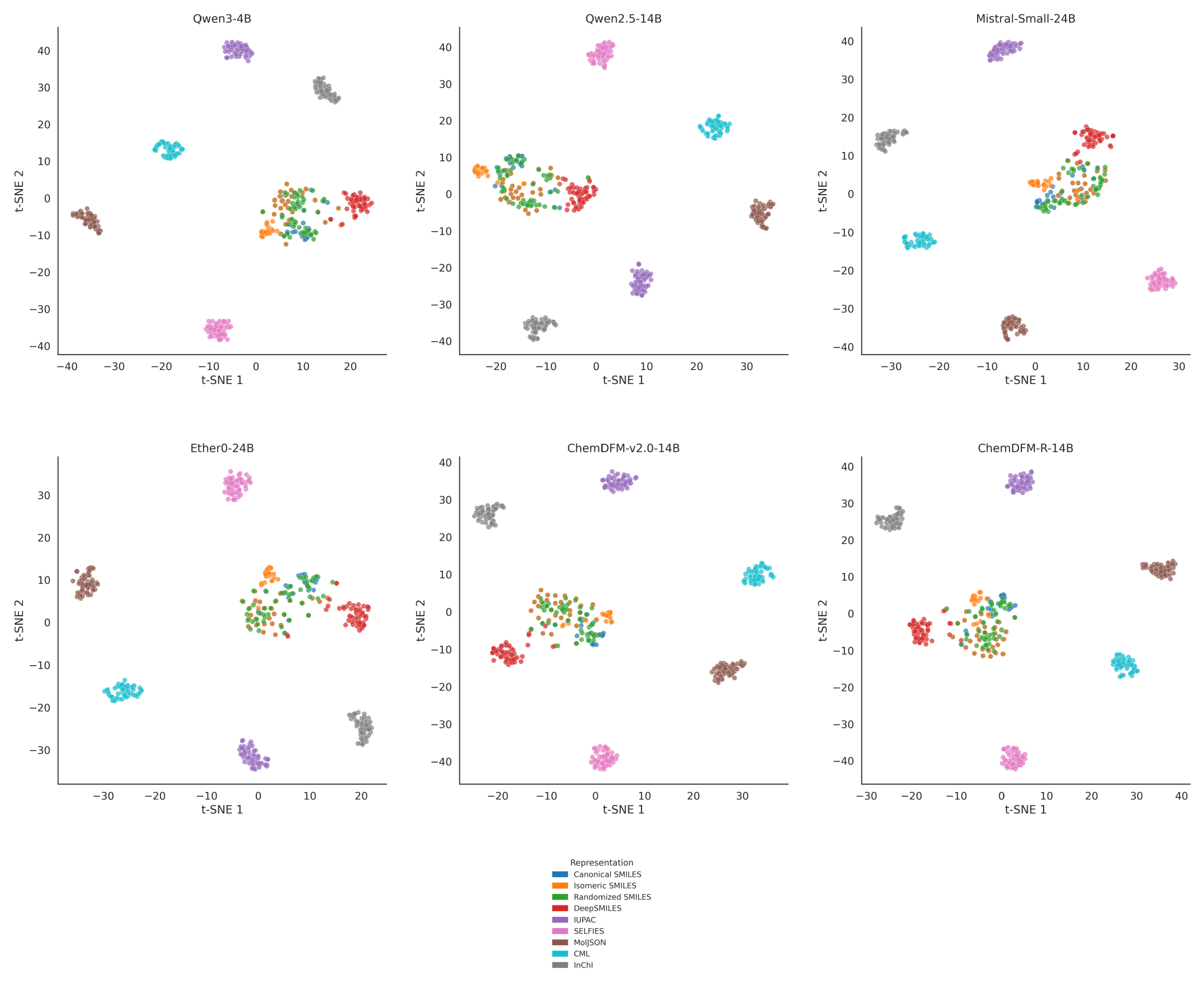}
  \caption{
t-SNE plots of the various text representations of 10 molecules with different LLMs   }
  \label{fig_tsne}
\end{figure}

\end{document}